\documentclass[lettersize,journal]{IEEEtran}
\usepackage{amsmath,amsfonts}
\usepackage[backend=biber,style=ieee]{biblatex}
\usepackage{subfloat}

\usepackage{algorithmic}
\usepackage{algorithm}
\usepackage{array}
\usepackage{subcaption}
\usepackage{textcomp}
\usepackage{stfloats}
\usepackage{url}
\usepackage{verbatim}
\usepackage{graphicx}
\usepackage{subcaption}
\usepackage{booktabs}
\usepackage{colortbl}
\usepackage{tikz}
\usepackage{url}
\usepackage{amssymb}
\usepackage{subcaption}
\usetikzlibrary{positioning, arrows.meta, shapes.geometric}
\begin{document}

\title{Multi-Agent Spectrum Sharing}

\author{Job Elliott,~\IEEEmembership{~Graduate Student Member, IEEE}, Justin G. Metcalf, ~\IEEEmembership{~Senior Member, IEEE}, Golnaz Habibi, ~\IEEEmembership{~Senior Member, IEEE}}

\markboth{IEEE TRANSACTIONS ON RADAR SYSTEMS,~Vol.~4,~2026}%
{Shell \MakeLowercase{\textit{et al.}}: A Sample Article Using IEEEtran.cls for IEEE Journals}

\IEEEpubid{}


\maketitle
\newcommand{\gh}[1]{\textcolor{blue}{G.H: #1}}
\newcommand{\je}[1]{\textcolor{red}{J.E: #1}}

\begin{abstract}
This project explores how multiple cognitive radars can learn to share limited wireless spectrum with other radio users without interfering with one another. Using machine learning (ML), each device independently decides where and how widely to transmit within a fixed 100~MHz band. The system analyzes real or simulated signal activity to detect which parts of the spectrum are currently in use and which are open. Based on this information, the devices adapt their transmission choices to avoid crowded frequencies while making efficient use of available space. The goal is to develop a flexible, scalable approach to spectrum sharing that could support future wireless communication systems.
Experimental results using both over-the-air software-defined radio (SDR) recordings and simulated environments demonstrate that the proposed meta-learning approach consistently balances competing objectives better than conventional reinforcement learning (RL) methods in multi-agent spectrum-sharing scenarios. 
Across five multi-agent benchmark environments, our proposed method achieved the highest average reward among the primary baseline algorithms while simultaneously maintaining low collision rates and stable transmission behavior.
\end{abstract}

\begin{IEEEkeywords}
Cognitive radar, spectrum sharing, multi-agent reinforcement learning, dynamic spectrum access, software-defined radio, continuous control, waveform adaptation, radar coexistence.
\end{IEEEkeywords}

\section{Introduction}
\IEEEPARstart{A}{ccess} to the electromagnetic spectrum is critical for radar, communications, and sensing systems. Each application requires large bandwidths for high quality-of-service, yet the spectrum is increasingly congested due to growth of commercial wireless networks and expansion of connected devices.

Future wireless environments will likely require radar, communications, and sensing systems to operate cooperatively in increasingly congested spectral regions. Traditional fixed-allocation spectrum policies are becoming impractical as the number of wireless devices and autonomous platforms continues to grow. Radar systems in particular require large bandwidths to achieve high range resolution and reliable target detection, yet these systems are often deployed in environments already occupied by commercial and unlicensed users.

Existing heuristic approaches such as sense-and-avoid (SAA) provide a simple mechanism for identifying open spectrum regions \cite{martone2018, kirk2018}, but they lack the ability to predict future occupancy patterns or coordinate behavior across multiple adaptive users. At the same time, many RL approaches focus primarily on single-agent adaptation and do not adequately address the non-stationary behavior introduced by multiple learning-enabled transmitters operating simultaneously \cite{flandermeyer2024, hausknecht2015}.



These challenges motivate a decentralized learning framework capable of balancing interference avoidance, spectral efficiency, and radar waveform stability in dynamic multi-agent environments. Existing heuristic methods cannot coordinate multiple adaptive users, while many reinforcement learning approaches either focus on single-agent settings or rely on discretized state and action spaces that limit spectrum resolution and waveform adaptation. Consequently, there remains a need for scalable continuous-control methods that enable multiple cognitive radars to learn cooperative spectrum-sharing behaviors directly from spectrum observations.

To address these limitations, this paper proposes a decentralized multi-agent reinforcement learning framework for cognitive radar spectrum sharing in which each radar independently learns continuous transmission parameters from local spectrum observations. The proposed approach extends model-free opponent shaping (M-FOS) by introducing a recurrent policy capable of adapting transmission center frequency, and bandwidth while simultaneously meta-learning policy hyperparameters during training. Unlike conventional proximal policy optimization (PPO) and deep-Q network (DQN) approaches, the proposed framework explicitly targets decentralized coordination among multiple adaptive radar agents operating in a shared spectrum without requiring centralized control or communication. Performance is evaluated using both real over-the-air SDR recordings and simulated spectrum environments under single-agent and multi-agent operating conditions.

\section{Related Work}
\subsection{Spectrum Sharing}
Spectrum sharing is a functionality that coordinates the spectrum usage among different users aiming at minimizing collisions and interference \cite{Tragos2013, ieeeDefinitions}.
One prominent example is the FCC Citizens Broadband Radio Service (CBRS), which introduced a three-tier spectrum access framework consisting of Incumbent Access, Priority Access Licenses (PAL), and General Authorized Access (GAA) for the 3.5~GHz band \cite{FCCCBRS, Voicu2019}.
First priority was given to Department of War radars (incumbents), then to commercial carriers who purchased temporary rights at auction (PAL), and finally open unlicensed access for the public (GAA). As demand grows for spectrum bandwidth, it becomes increasingly likely that incumbent users may be forced to share their prioritization and work around other users. It is also a similar problem in hostile spaces where jamming is a concern \cite{jiang2022}.

\subsection{Reinforcement Learning}
Reinforcement learning is a machine learning paradigm in which an agent learns a decision-making policy by interacting with an environment and receiving scalar reward feedback. The objective is to learn a policy that maximizes the expected cumulative reward through repeated trial-and-error interactions, balancing exploration of new actions with exploitation of previously learned behaviors \cite{Sutton2018}.
PPO is a popular on-policy RL algorithm introduced in 2017 by OpenAI which achieved stable, smooth updates by preventing aggressive sudden policy changes over single updates \cite{schulman2017}. It has been applied in a wide range of applications including game theory \cite{lu2022} and, more pertinently, spectrum sharing \cite{flandermeyer2024}. DQNs were introduced by \cite{mnih2015}, and they have been applied to the spectrum sharing problem in \cite{hausknecht2015}, and this method was compared with the PPO algorithm in \cite{flandermeyer2024}.

\subsection{Cognitive Radio / Cognitive Radar}
Cognitive Radio is a concept formally published in 1999 by Joseph Mitola \cite{mitola1999}. It involves radio devices which are aware of their environment and make transmission decisions based on their observations of that environment \cite{haykin2005}. A cognitive radar is simply an application of cognitive radio for detection and ranging purposes. Cognitive radar uses a closed-loop perception-action framework in which the radar senses the electromagnetic environment and adapts waveform parameters to mitigate interference while maintaining sensing performance. In this process, transmissions are received by the radar, those transmissions are processed and used as input for reinforcement learning to train the cognitive radar's policy which then outputs the next transmission based on predictions on the environments following states.


\subsection{Multi-Agent Reinforcement Learning and Meta-Learning}

Multi-agent reinforcement learning (MARL) has been explored in game theory scenarios \cite{foerster2018, lu2022}, vehicular spectrum sharing for communications \cite{liang2019spectrum}, distributed cognitive radio scenarios \cite{tao2026}, and in multiple-input and multiple-output (MIMO), phased-array, and digital array radars \cite{dolinger2025}. We apply meta-learning concepts from model-free opponent shaping (M-FOS) \cite{lu2022} which demonstrated strong performance in multi-agent competitive environments similar to the environment we envision, in order to create our own M-FOS style agent. Meta-learning in our case involves manipulating the hyperparameters of an inner neural network learning policy. This can help automate the tuning of these parameters, which is especially valuable if the ideal values for those parameters is dynamic within a single environment or across different environments.

The scenario considered in this work involves multiple independent cognitive radar agents operating within a shared spectral environment alongside non-cognitive users. We investigate both shared and independent learning strategies and evaluate whether meta-learning techniques can improve coordination and spectrum-sharing performance in these dynamic multi-agent settings.

\subsection{Problem Statement}

This work considers the problem of distributed spectrum sharing among multiple cognitive radar agents operating within a shared 100~MHz spectral environment containing a variety of emitter types. Each radar agent independently observes spectrum occupancy and selects continuous transmission parameters consisting of center frequency and bandwidth. The objective of each agent is to maximize collision-free bandwidth utilization while minimizing interference with other spectrum users and maintaining waveform stability across coherent processing intervals (CPIs).

Unlike traditional single-agent reinforcement learning problems, the environment considered here is inherently non-stationary because each agent’s transmission decisions alter the spectral conditions experienced by the other agents. This creates a coupled multi-agent optimization problem in which locally optimal transmission strategies may produce globally inefficient spectrum usage patterns.

The problem is further complicated by radar-specific operational constraints. Excessive changes in center frequency or bandwidth between pulses can distort range-Doppler processing and degrade radar performance, even if spectrum collisions are avoided. Therefore, effective spectrum sharing requires balancing multiple competing objectives including interference mitigation, spectral efficiency, waveform consistency, and adaptability to dynamic occupancy conditions.

This work formulates the spectrum sharing problem as a cooperative multi-agent continuous-control reinforcement learning task in which agents learn decentralized transmission policies directly from spectrum observations and environmental feedback.

\subsection{Contributions}
This work makes several contributions to the study of cognitive radar spectrum sharing. 
\begin{itemize}
\item First, the spectrum-sharing problem is formulated as a decentralized MARL task with continuous action spaces, allowing agents to jointly adapt their transmission parameters in dynamic spectral environments. 

\item Second, an M-FOS meta-learning architecture is introduced that combines convolutional neural network (CNN) encoding with a Gated Recurrent Unit (GRU) for the policy networks with genetic-algorithm-driven hyperparameter adaptation to improve learning performance across varying operating conditions. 

\item Third, the proposed approach is evaluated using both over-the-air SDR datasets and simulated spectrum environments, providing insight into performance under realistic and controlled conditions. 

\item Finally, the work investigates spectrum coordination behavior in both single-agent and multi-agent settings, enabling a comparative analysis of agent interactions and spectrum-sharing effectiveness.
\end{itemize}

\section{Problem Formulation}
This problem is formulated as a partially observable Markov decision process (POMDP) in which radar agents observe spectrum occupancy and select frequency allocations to maximize long-term spectrum utilization while minimizing interference.

\subsection{Spectrum Sharing Environment}
The spectrum sharing environment models a congested 100~MHz radio-frequency band represented in the frequency domain using a Fast Fourier Transform (FFT) based discretization of size $N = 1024$. 
Each agent observes the occupancy of the entire 100 MHz spectrum from its own perspective, with its own transmission excluded from the observation. Although the observation spans the full spectrum, agents do not have information identifying the sources of occupied regions or the actions or policies of other agents. Thus, each agent has an agent-specific observation of the global spectrum state despite observing the full frequency range.
Spectral occupancy is represented as a sequence of snapshots sampled every 10.24 $\mu$sec, corresponding to the temporal resolution of both the recorded and simulated datasets. Recorded and simulated emissions are modeled as incumbent or priority access license (primary) users, while the cognitive radar agents operate as general authorization access (secondary) users and must avoid interfering with incumbent transmissions. Each agent selects an action every pulse which is comprised of a 10.24 $\mu$sec pulse width (PW) with a 204.8 $\mu$sec pulse repetition interval (PRI) which gives a 1:19 ratio of transmission to observation time. At each pulse, an agent selects a spectral allocation defined by a center frequency and bandwidth, which is mapped to the corresponding FFT bins. A collision occurs whenever an FFT bin is simultaneously occupied by multiple cognitive radar agents or by both a cognitive radar agent and a primary user. The total number of overlapping bins is used to compute the collision penalty associated with each action.

\subsection{Pulse-Doppler Radar Constraints}

Unlike conventional communication systems, cognitive radar waveform adaptation must preserve coherence across a CPI. Pulse-Doppler radar estimates target velocity by coherently processing multiple transmitted pulses, and this processing assumes that waveform characteristics remain relatively stable throughout the CPI \cite{richards2014}.

Frequent changes in center frequency or bandwidth can violate this assumption and introduce range sidelobe modulation (RSM), which distorts the resulting range-Doppler response \cite{higgins2011}. Consequently, a spectrum-sharing policy that aggressively changes transmission parameters may achieve low interference while simultaneously degrading radar performance.

To account for this tradeoff, the proposed reward function includes penalties for pulse-to-pulse changes in center frequency and bandwidth. Range-Doppler responses are later used as a qualitative measure of waveform stability and radar coherence under different spectrum-sharing strategies in Fig.~\ref{fig:rngVelPlots} in a similar manner performed in \cite{flandermeyer2024}.

\subsection{State and Action Space}

Each agent observes the spectrum using a fast spectrum sensing (FSS) energy detector \cite{martone2018}, which produces a binary occupancy representation
$x_t \in \{0,1\}^{N_{FFT}}$ for each snapshot. The detection threshold is adaptively computed using the hardware-optimized cell averaging estimation (HO-CAE) method in \cite{mattingly2023}, improving robustness under varying noise conditions.

Following \cite{flandermeyer2024}, we construct the observation space as a temporal stack of the most recent $N_{snap}$ snapshots, forming an
$N_{snap} \times N_{FFT}$ binary occupancy matrix that preserves both spectral and temporal structure.

Actions are parameterized by center frequency and bandwidth and mapped to FFT-bin allocations within the monitored spectrum. DQN agents operate on a discretized version of this action space, while PPO and M-FOS agents operate directly on the continuous parameterization. 

\subsection{Multi-Agent Setup}

{In our experiments, the problem is formulated in a cooperative manner, the desire is to neither interfere with other agents, nor be interfered with, which partially accounts for malicious strategies as well (such as jamming, though this is not explored in this work). Each agent is only seeking to maximize its reward, which is essentially a greedy approach, but the penalties for interfering with others allows the different agents to find mutually beneficial policies that neither crowd out the other agents, nor allow themselves to be crowded out.}

\subsection{Reward Definition}
Our reward function is similarly structured to \cite{flandermeyer2024} where transmission bandwidth is positively rewarded, collision/interference is heavily penalized, and bandwidth and center frequency distortion are penalized moderately. 
The proposed reward function differs from prior work in four primary ways:
Agents are allowed to choose actions that go beyond the 100~MHz bandwidth; however, any transmission out of bounds is considered a collision. 
Only clean (non-colliding) transmission is rewarded. 
There is no negative penalty for the transmission bandwidth being smaller than the largest possible available bandwidth (mostly to prevent encouraging the agent to fill gaps between other agent's pulses).
Collisions are computed not just based on two active transmissions in the same frequency, but are based on spectrum ownership so that if an agent is sending pulses, and a second agent intrudes by sending a pulse between the first agent's pulses, the second agent is penalized for interference and the first agent receives no penalty.
Lastly, an additional small collision penalty is added for the duration that an agent is in the same frequency as another agent but is only listening. This is because it is detrimental to the agent to have another agent interfering with its PRI, but it is not harmful to the intruding agent. This encourages the agent to find areas not only where it does not interfere with others, but also to find areas where it is not interfered with by others. In our experiments, this penalty is a 30th of the typical collision penalty, however, the total reward for a PRI is summed so that the agents reward per decision is made up of the active reward and the passive reward.

Our reward function can be formulated as follows:

\begin{equation}
R_{total} = (\sum{R_{active}}) + (\sum{R_{passive}}) - R_{adapt}
\end{equation}

\begin{equation}
R_{active} = B_r - \alpha_c B_c
\end{equation}

\begin{figure*}[t]
\centering
\includegraphics[width=1\linewidth]{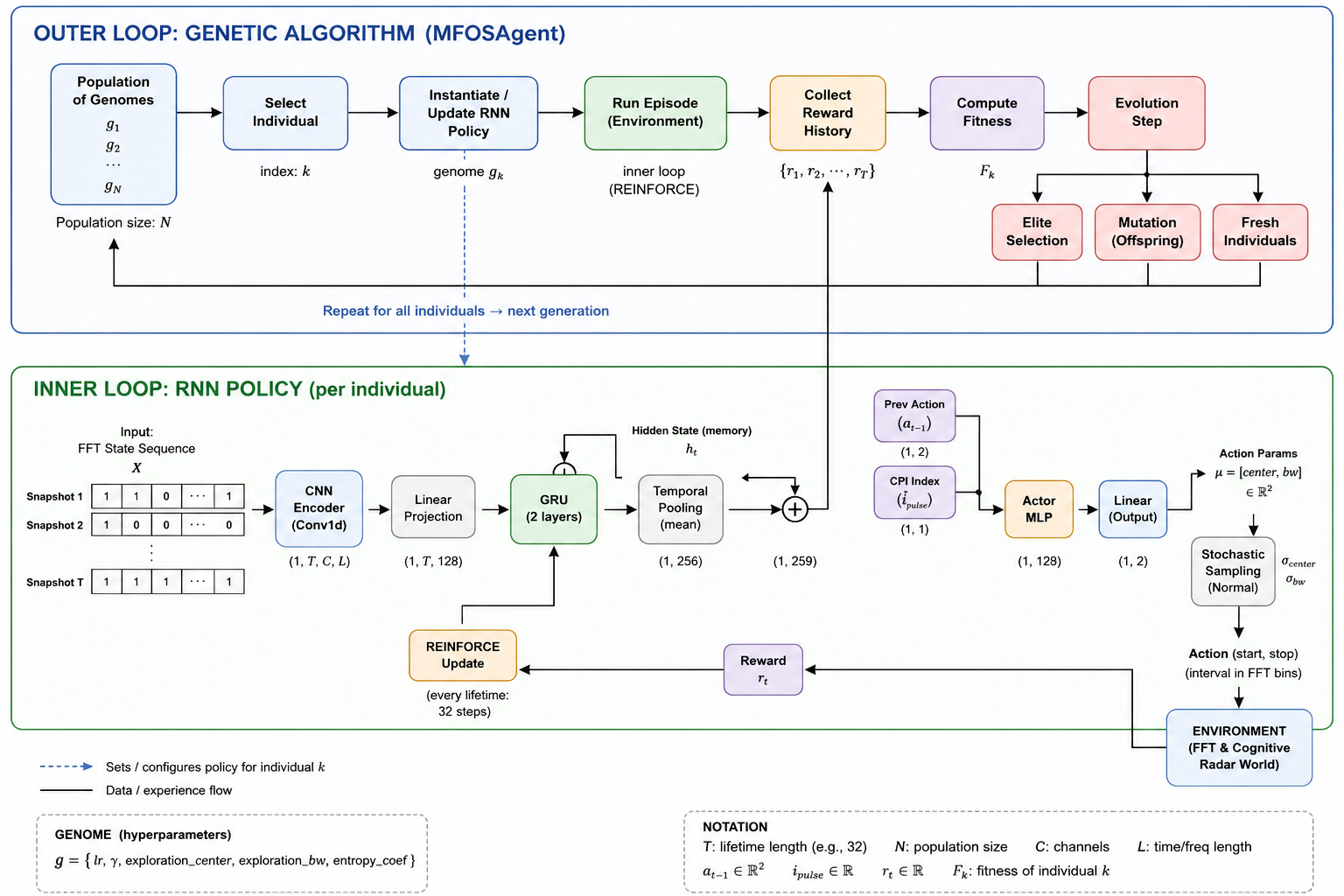}
\caption{M-FOS Architecture}
\label{fig:M-FOSArchitecture}
\end{figure*}

\begin{equation}
\label{equ:passive}
R_{passive} = - \frac{\alpha_c B_c}{30}
\end{equation}

\begin{equation}
\label{equ:adaptation}
R_{adapt} = \beta_{bw}|B_r - \mu_{B_r}| + \beta_{f_c}|f_c - \mu_{f_c}|
\end{equation}

Where $R_{active}$ is the reward for an active transmission, $R_{passive}$ is the reward for the non-transmitting portion of each PRI, and $R_{adapt}$ is the adaptation penalty for changing the bandwidth and center frequency within a CPI. $B_r$ is the total collision-free transmitted bandwidth, $\alpha_c$ is the collision weight, and $B_c$ is the amount of bandwidth that collides with any other transmission (or spectrum reserved by another radar between pulses). The passive reward in (\ref{equ:passive}) discourages an agent from dwelling in regions occupied by other users while it is not transmitting. Because each PRI contains 19 passive observations but only a single active transmission opportunity (see the discussion below), the passive penalty is divided by 30. This scaling prevents the accumulated passive rewards from dominating the total return while ensuring that the active transmission reward remains the primary driver of the learned transmission policy. For the adaptation equation (\ref{equ:adaptation}), $\beta_{bw}$ and $\beta_{f_c}$ are tunable weights for the bandwidth distortion and the center frequency distortion respectively. $|\cdot|$ is the absolute value operator, $\mu_{B_r}$ is the average bandwidth for the radar over the current CPI, $\mu_{f_c}$ is the average center frequency over that same interval, and $f_c$ is the active center frequency.
In our experiments, the PW is 10.24 $\mu$sec which corresponds to a single spectrum sample and the PRI is 204.8 $\mu$sec which is equivalent to 20 samples. This means our transmission to listening ratio is $1:19$.

The reward function is intended to approximate the competing objectives faced by practical cognitive radar systems rather than optimize a single metric. Specifically, the reward balances three operational considerations: maximizing usable collision-free bandwidth, minimizing interference with other spectrum users, and maintaining waveform consistency across a CPI. Optimizing any one of these objectives in isolation can produce undesirable behavior. For example, maximizing bandwidth alone may increase interference, while minimizing waveform distortion alone may discourage adaptation to changing spectral conditions. The reward formulation therefore serves as a scalar approximation of a multi-objective optimization problem encountered in practical spectrum-sharing environments.

\section{Methodology}

\subsection{Agent Architectures}
We propose a meta-learning agent based on M-FOS which consists of an inner policy made up primarily of a CNN and GRU. The CNN is used to encode the information from the $N \times FFT$ frequency occupancy snapshots and feed that information to the GRU, which helps to maintain the hidden state across timesteps. The outer policy is an application of the genetic algorithm (GA) which is used to fine-tune the learning rate $\alpha$, discount factor $\gamma$, entropy coefficient, exploration center, and exploration bandwidth. Ablation experiments were performed by removing the outer GA which acts as the meta-learning part of the M-FOS agent. The model architecture is shown in Fig.~\ref{fig:M-FOSArchitecture}.

\subsection{Training Procedure}

For each SDR dataset, the recorded spectrum occupancy data were partitioned into non-overlapping training (80\%) and evaluation (20\%) segments. During training, agents interacted only with environments constructed from the training portion of the recordings. All reported results were generated using environments constructed from the held-out evaluation portion to measure generalization to previously unseen spectrum activity.

During training, agents interacted continuously with spectrum environments constructed from prerecorded and simulated RF datasets. In the multi-agent setting, heterogeneous agents (PPO, DQN, M-FOS, and ablated  M-FOS) operated concurrently within the same environment, each maintaining an independent policy and learning process but sharing the same reward function.

All agents operated within a shared spectrum environment but were not synchronized. Each radar selected a new transmission action once every pulse repetition interval (PRI), corresponding to every 20 spectrum snapshots. Since the transmitted pulse occupied only a single snapshot within each PRI, agents spent the remaining snapshots in a passive listening state. To better reflect independent real-world radar systems, each agent was initialized with a random offset within the 20-snapshot PRI, preventing all agents from transmitting and listening simultaneously. Following each transmission, the occupied bandwidth remained marked as reserved throughout the subsequent listening period, allowing other agents to incur interference penalties even when pulses did not overlap in time. Consequently, collisions between learning agents most often resulted from transmitting into spectrum that had been reserved by another radar during its most recent pulse rather than from simultaneous pulse overlap. Agent actions were also evaluated against the prerecorded or simulated emitters present in the environment, allowing interference to occur with both learning agents and background RF activity. No communication or parameter sharing occurred between agents, and each maintained an independent policy, optimizer, replay buffer (when applicable), and learning process. Interaction therefore occurred only indirectly through the shared RF environment and the resulting reward signal.

For evaluation, deterministic policy execution was used whenever applicable. PPO, DQN, and the inner policy of the M-FOS agents selected actions without stochastic exploration at test time. The M-FOS agents were evaluated using the best-performing genome identified at the end of training which during evaluation only involves the exploration bandwidth and exploration center frequency parameters.

For transfer learning experiments, pretrained models obtained from one spectral environment were used to initialize agents before further training or evaluation on a different dataset. This setup was used to assess generalization of learned spectrum-access strategies across varying frequency bands and occupancy conditions.

\subsection{Proposed Method}


M-FOS: This paper proposes a meta-learning model based on Model-free opponent shaping \cite{lu2022} using a CNN-GRU shared inner policy with a genetic algorithm (GA) 
outer policy to fine-tune the hyperparameters. 
M-FOS is trained in a decentralized multi-agent spectrum-sharing environment in which each radar independently learns a transmission policy from its local spectrum observations while interacting with other agents through a shared RF environment. No communication or parameter sharing occurs between agents, and coordination emerges solely through the shared reward function and spectrum contention.
The authors of \cite{lu2022} experiment with both PPO and GA for meta-learning. GA was selected for this paper for its lack of a gradient update step which allows it to be much quicker in processing. However, PPO may potentially learn a good policy in fewer iterations that GA, so GA may or may not be optimal for this task. No comparison of these different meta-learning approaches is made here.
The population-based GA was used to evolve high-level hyperparameters governing the agent’s learning and exploration behavior. At each evolutionary cycle, candidate genomes were evaluated in the environment using cumulative reward, after which elite genomes were retained, mutated, and supplemented with randomly initialized genomes to preserve diversity.
All genomes shared a single underlying policy network, consisting of a CNN-based encoder followed by a GRU-based temporal model. This policy was trained using standard gradient-based optimization driven by the environment reward signal, while the evolutionary process operated on the hyperparameters controlling its behavior. This separation allows fast adaptation of learning dynamics through evolution while maintaining a consistent shared representation learned by the policy network.

The action selection space was identical to the PPO implementations: any continuous center frequency selection and any continuous bandwidth selection with the same 102 FFT bins (approximately 10~MHz) minimum.

\begin{figure*}[htbp]
    \centering
    \subfloat[ARRC 2.4--2.5~GHz
    \label{fig:245arrcSample}]{
        \includegraphics[width=.48\linewidth]{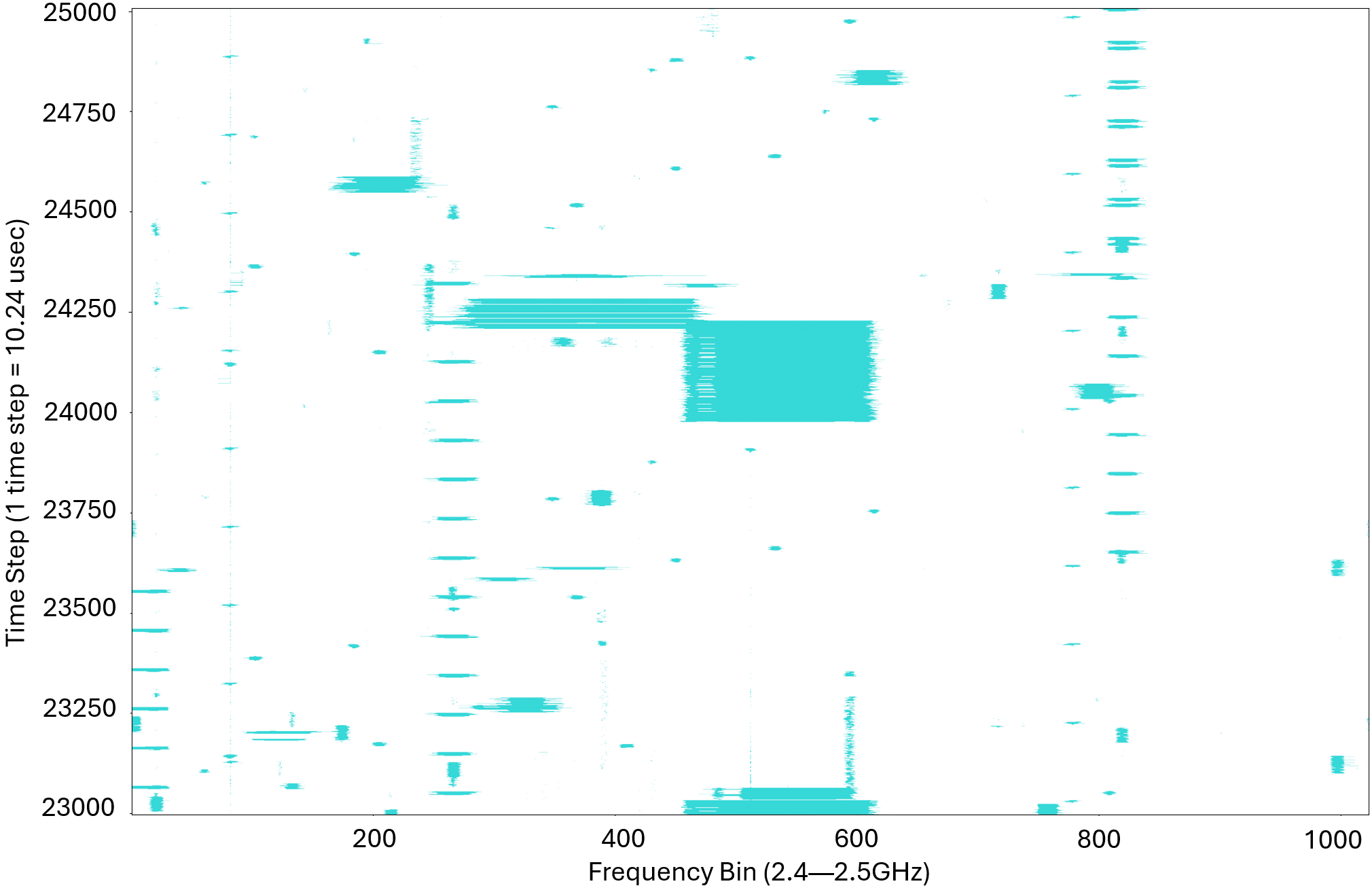}
    }
    \hfill
    \subfloat[Oklahoma Memorial Union 2.4--2.5~GHz
    \label{fig:245unionSample}]{
        \includegraphics[width=.48\linewidth]{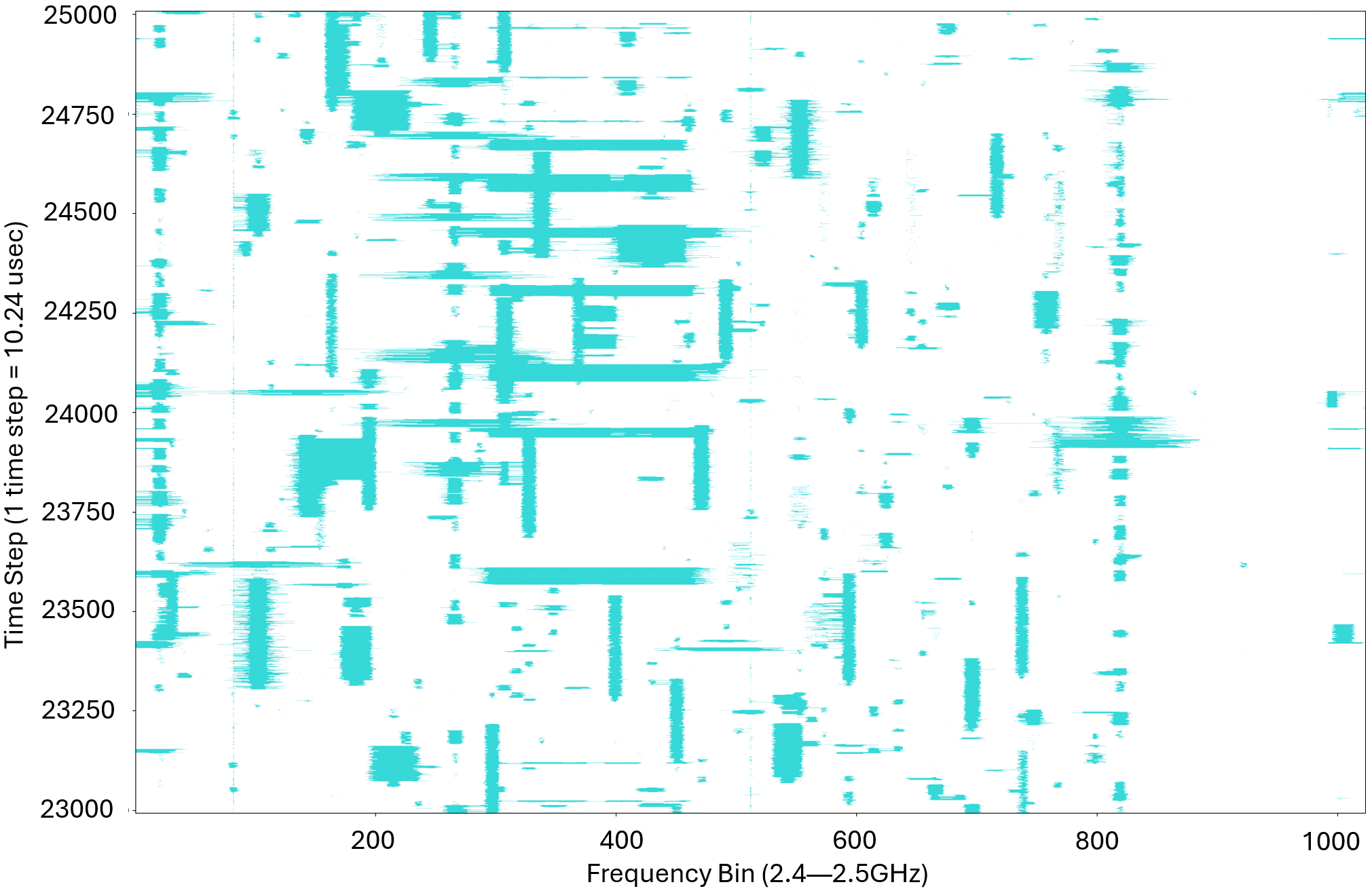}
    }
    \hfill
    \subfloat[ARRC 2.59--2.69~GHz
    \label{fig:264aarcSample}]{
        \includegraphics[width=.48\linewidth]{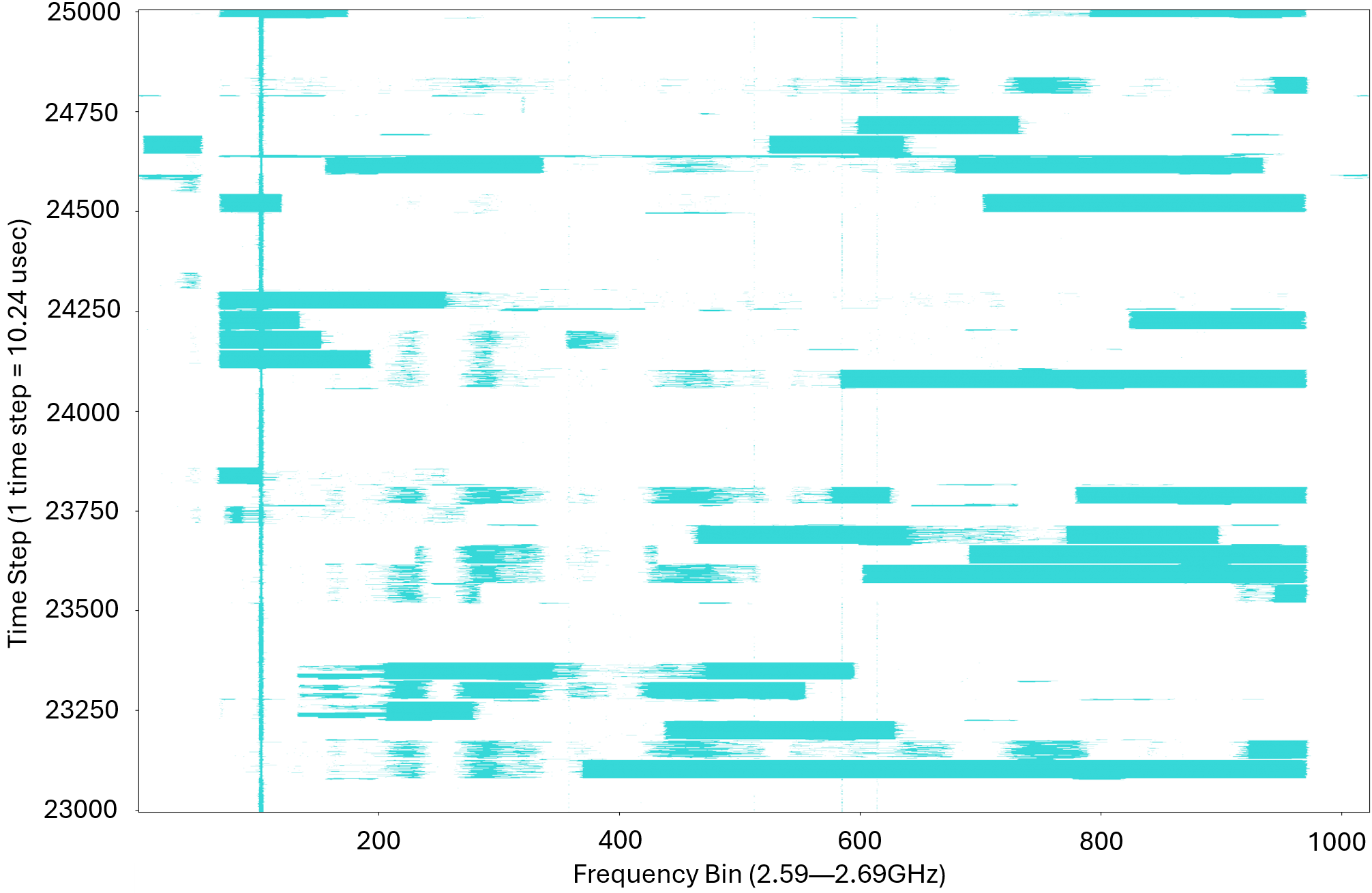}
    }
    \hfill
    \subfloat[Oklahoma Memorial Union 2.59--2.69~GHz
    \label{fig:264unionSample}]{
        \includegraphics[width=.48\linewidth]{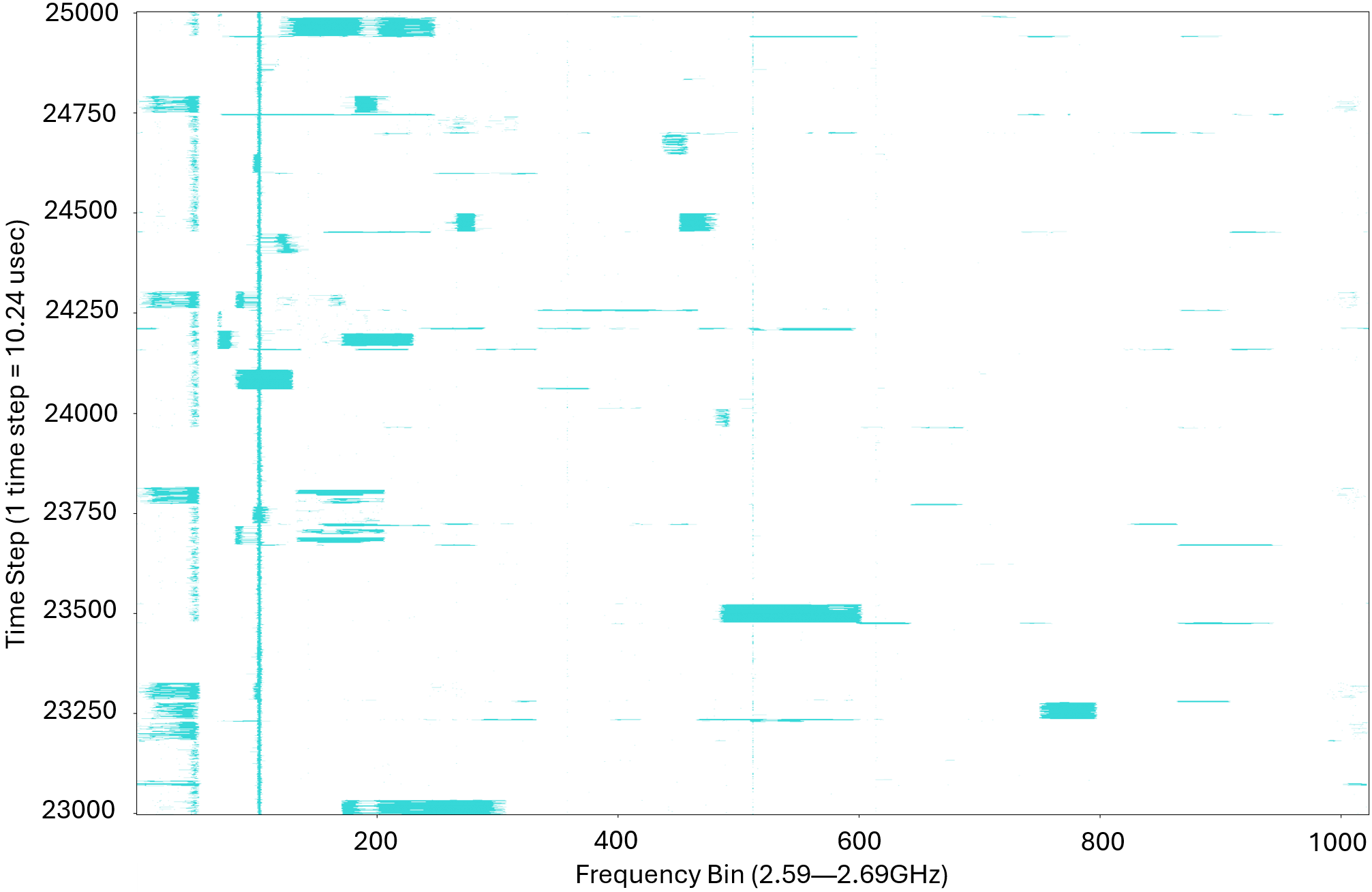}
    }
    \caption{Collected 0.02 Second Spectrum Data Samples}
    \label{fig:spectrumSamples}
\end{figure*}

\section{Dataset}
\subsection{SDR Datasets}
Over-the-air spectrum data were collected using an Ettus Research USRP X310 equipped with UBX-160 daughterboards, controlled through GNU Radio and UHD drivers. Recordings were obtained in two campus buildings, the Advanced Radar Research Center (ARRC) and the Oklahoma Memorial Union, within the 2.4--2.5~GHz and 2.59--2.69~GHz frequency bands. Each recording captured a 100~MHz observation bandwidth used for spectrum sensing and occupancy analysis.

The collected datasets contain a diverse set of non-cooperative emitters exhibiting a wide range of spectrum occupancy patterns, congestion levels, and transmission behaviors. These variations provide representative real-world operating conditions for evaluating adaptive spectrum-sharing agents. Samples of the recorded datasets are shown in Fig.~\ref{fig:spectrumSamples}, illustrating the diversity in spectrum congestion, emission patterns, and occupied frequencies across the collection environments.

\subsection{Simulation Environment}

Simulated data were generated to supplement the recorded datasets for training and evaluation. The simulator models four representative emitter classes—skinny, wide, pulsed, and rectangular—operating within the 100 MHz observation band. For each emitter, a random center frequency was selected uniformly across the observation band while ensuring that the entire signal bandwidth remained within the observable spectrum. Bandwidths and temporal characteristics were sampled uniformly from predefined ranges summarized in Table \ref{tab:simulation_parameters}. 

\begin{table}[htbp]
\caption{Simulation emitter parameters.}
\label{tab:simulation_parameters}
\centering
\renewcommand{\arraystretch}{1.1}
\begin{tabular}{l|c|p{2.2cm}|p{2.3cm}}
\toprule
\textbf{Type} &
\textbf{BW (MHz)} &
\textbf{Tx / Idle} &
\textbf{Action Update} \\
\midrule
Skinny &
1.46--2.93 &
2.3--3.58 ms / 19.46--21.50 ms &
Every 307.2 ms \\

Wide &
14.65--17.09 &
1.02--2.56 ms / 92.16--102.40 ms &
Every 1024 ms \\

Pulsed &
2.44--4.88 &
0.21--0.51 ms / 0.51--0.82 ms (within 20.48--51.2 ms burst cycles) &
Every 1024 ms \\

Rectangular &
5.86--66.41 &
0.46--0.56 ms / 0.1 ms--2.56 ms &
Every Pulse \\
\bottomrule
\end{tabular}
\end{table}

\begin{figure}[htbp]
\centering
\includegraphics[width=1\linewidth]{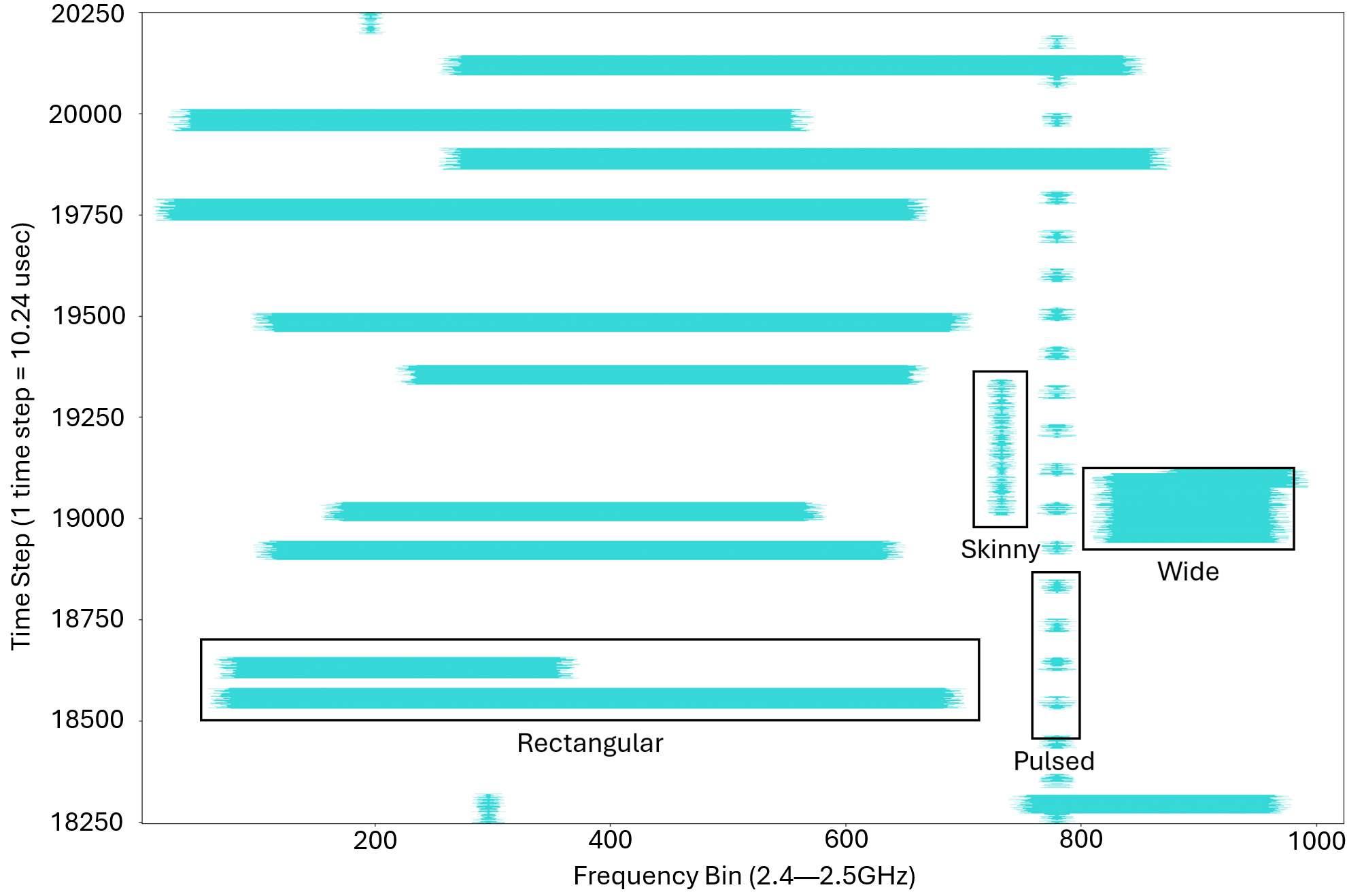}
\caption{Simulation Data .02 Second Sample}
\label{fig:simSample}
\end{figure}

To emulate the small frequency drift observed in measured recordings, each emitter was also subjected to a random frequency offset of up to approximately 2 MHz between successive snapshots. These parameter ranges were chosen to approximate transmission behaviors manually observed in the recorded datasets while providing sufficient diversity for reinforcement learning. The simulator enables the generation of arbitrarily large training and evaluation datasets with configurable levels of spectrum congestion without requiring additional data collection. Although simulated environments cannot reproduce every characteristic of real RF environments, they provide a repeatable platform for algorithm development, and performance obtained using simulated data should therefore be interpreted as complementary to results on measured data. A sample of the simulated environment is shown in Fig.~\ref{fig:simSample}.

The implementation and processed datasets are publicly available in~\cite{yourrepo2026}.

\section{Experimental Setup}

\subsection{Baselines}
The proposed approach is evaluated against four baseline methods representing a range of adaptive and non-adaptive spectrum-sharing strategies.

The first baseline is SAA, a standard sense-and-avoid heuristic. SAA considers only the most recent spectrum snapshot (spanning a 10.24 $\mu$sec interval in our setup) and selects the largest contiguous region of unoccupied frequency bins for transmission. Because the algorithm operates solely on the current observation, it has no concept of spectrum occupancy persistence or bin ownership between pulses.

The second baseline is a Random Static Action policy, which serves as a non-adaptive reference. At the start of an experiment, the policy randomly selects a bandwidth between 10 and 30~MHz and a center frequency within the 100~MHz operating range such that the selected bandwidth remains entirely within the available spectrum. The selected action remains fixed for the duration of the experiment and is never modified in response to environmental conditions.

The third baseline is a PPO agent with continuous action control \cite{flandermeyer2024}. Although a Multi-Agent PPO (MAPPO) implementation was initially considered, the fully observable nature of the spectrum occupancy environment limits the benefit of a centralized critic, as it would primarily provide information regarding which specific agent occupies a given portion of the spectrum. Consequently, the PPO implementation from \cite{flandermeyer2024} was adopted with a few modifications. During evaluation, action sampling is disabled to reduce center-frequency and bandwidth variability while preserving exploration during training. Additionally, actions that extend beyond the 100~MHz spectrum bounds are not artificially shifted into the valid action space. Instead, such out-of-bounds selections are treated as collisions and penalized through the reward function. The PPO agents' bandwidth selection was limited to a minimum of 102 FFT bins (approximately 10~MHz).

The fourth baseline is a DQN agent operating on a discretized action space \cite{wang2018}. The implementation largely follows the approach described in \cite{flandermeyer2024}. However, to provide a more equitable comparison with PPO, which operated with a minimum bandwidth of 10~MHz, the available DQN bandwidth selections were modified from approximately 3, 6, and 10~MHz to 10, 13, and 16~MHz.

Finally, an ablated version of our proposed M-FOS agent where there is no fine tuning of the hyperparameters with genetic algorithm is also employed to provide comparison to see if the meta-learning improves the performance meaningfully.

\subsection{Evaluation Metrics}
We record a number of metrics that can be compared. Average reward ($\bar{R}$) is the primary metric because it corresponds directly to the objective optimized during training. Superior models, whether due to architectural differences, meta-learning strategies, or continuous action spaces, should achieve higher average rewards, while a lower reward standard deviation ($\sigma_R$) indicates greater policy stability when accompanied by a high average reward. Reward is not the only meaningful metric though. We also record the average and standard deviation of collision bandwidth ($\bar{C}$ and $\sigma_C$), transmitted bandwidth (BW and $\sigma_{BW}$), bandwidth variation between consecutive transmissions ($\Delta BW$ and $\sigma_{\Delta BW}$), and center-frequency variation between consecutive transmissions ($\Delta CF$ and $\sigma_{\Delta CF}$). If any single metric is optimized to the neglect of the others, this would reveal an issue with the agent being able to balance the different tasks. Better balance can likely be achieved for each model individually by fine-tuning the reward function, but in all our experiments the same reward function is shared by all agents for standardization.

In addition to these quantitative metrics, we also compare convergence rates and qualitative behavior observed in the learning curves, spectrum occupancy plots, and range-Doppler plots. For example, the spectrum occupancy plots reveal failures to adapt transmissions to avoid collisions or policies that converge to nearly static transmission strategies. The range-Doppler plots illustrate waveform distortion resulting from bandwidth and center-frequency variations within a CPI, although they do not account for interference effects.

\subsection{Solo-Agent Experiments}

Solo experiments were conducted where each agent runs independently of any other agents inside of five different datasets. In each environment, the experiment was run three times with different random seeds and starting indices for each agent type, and the average results for each of the three runs per agent were recorded. The simulation environments were generated using a configuration of three 'wide,' four 'skinny,' and five 'pulsed' emitters, emulating the congestion amount observed in the 2.4--2.5~GHz ARRC dataset.

\subsection{Multi-Agent Experiments}

The experiment with results shown in Table~\ref{tab:multiTotalRewards} was run with five agents simultaneously interacting, and the total rewards summed over the evaluation period for each agent. The other experiments, Table~\ref{tab:multiResults}, involved running three agents simultaneously, and taking the average of all metrics for the three agents in each experiment. Those tables also show the results of running a PPO, a DQN, and an M-FOS agent together, and averaging the results, this was conducted twice with different random starting seeds and the average of averages across both experiments was taken. 

\section{Results}
This section presents the experimental evaluation of the proposed framework in three stages. First, the solo-agent experiments establish baseline performance by evaluating each agent independently in recorded and simulated RF environments. Second, the multi-agent experiments investigate the primary contribution of this work by evaluating how multiple adaptive cognitive radar systems share spectrum without explicit communication or centralized coordination. Finally, transfer-learning experiments examine whether policies learned in one RF environment generalize to previously unseen spectrum conditions.
\subsection{Solo-Agent Results}
M-FOS achieves competitive performance across all individual-agent evaluation scenarios, while the heuristic SAA baseline obtains the highest average reward in every solo environment, as shown in Table~\ref{tab:indResults}. Among the learning-based approaches, M-FOS achieves the highest average reward in three of the five datasets, with DQN outperforming the remaining methods on the 2.59--2.69~GHz ARRC dataset in Table \ref{tab:ind264Arrc} and PPO achieving the highest reward on the 2.59--2.69~GHz Memorial Union dataset in Table \ref{tab:ind264union}. 

\begin{table*}[htbp]
\centering
\caption{ 
Performance comparison of all methods. Bold font indicates best value for that column while underlined numbers indicate the best value among the cognitive agents excluding the Random Static and SAA agents. Each value represents the average value per agent across three independent runs.}
\label{tab:indResults}

\begin{subtable}{\textwidth}
\caption{2.4--2.5~GHz ARRC Data, 1M Samples, Solo Average Performance}
\label{tab:ind245Arrc}
\centering
\small
\begin{tabular}{lcccccccccc}
\toprule
Agent & $\bar{R}\uparrow$ & $\sigma_R\downarrow$ & $\bar{C}\downarrow$ & $\sigma_C\downarrow$ & $\overline{\mathrm{BW}}\uparrow$ & $\sigma_{BW}\downarrow$ & $\Delta BW\downarrow$ & $\sigma_{\Delta BW}\downarrow$ & $\Delta CF\downarrow$ & $\sigma_{\Delta CF}\downarrow$ \\
\midrule
Random Static & -0.3619 & 1.4937 & 1.1006 & 2.6525 & 24.3555 & \textbf{0} & \textbf{0} & \textbf{0} & \textbf{0} & \textbf{0} \\
SAA & \textbf{0.1624} & \textbf{0.9070} & \textbf{0.2152} & \textbf{0.6974} & \textbf{61.1366} & 20.4962 & 17.4942 & 15.4910 & 15.4333 & 15.6210 \\
\midrule
PPO & -0.8621 & 2.1805 & 2.7534 & 4.2568 & \underline{60.1883} & 4.1026 & 1.5203 & 1.7911 & 1.7880 & 1.9204 \\
DQN & -0.2954 & 1.0093 & 0.4892 & 1.4126 & 12.3968 & 2.6035 & 2.3513 & 2.4659 & 24.6913 & 23.2452 \\
M-FOS & \underline{-0.1063} & \underline{0.9870} & \underline{0.4052} & \underline{1.0254} & 16.7513 & \underline{1.2691} & \underline{0.6218} & \underline{0.7685} & \underline{1.2682} & \underline{1.5971} \\
Abl. M-FOS & -1.5782 & 9.6904 & 1.0803 & 2.3716 & 24.4723 & 4.4498 & 2.7434 & 3.4369 & 2.4764 & 2.7137 \\
\bottomrule
\end{tabular}
\end{subtable}

\begin{subtable}{\textwidth}
\caption{2.59--2.69~GHz ARRC Data, 1M Samples, Solo Average Performance}
\label{tab:ind264Arrc}
\centering
\small
\begin{tabular}{lcccccccccc}
\toprule
Agent & $\bar{R}\uparrow$ & $\sigma_R\downarrow$ & $\bar{C}\downarrow$ & $\sigma_C\downarrow$ & $\overline{\mathrm{BW}}\uparrow$ & $\sigma_{BW}\downarrow$ & $\Delta BW\downarrow$ & $\sigma_{\Delta BW}\downarrow$ & $\Delta CF\downarrow$ & $\sigma_{\Delta CF}\downarrow$ \\
\midrule
Random Static & -1.3260 & 3.4185 & 3.0039 & 6.8790 & 24.3555 & \textbf{0} & \textbf{0} & \textbf{0} & \textbf{0} & \textbf{0} \\
SAA & \textbf{-0.0644} & \textbf{1.6162} & \textbf{0.2343} & \textbf{1.6516} & 54.0432 & 22.9701 & 18.5232 & 19.6341 & 16.5530 & 17.0048 \\
\midrule
PPO & -3.0612 & 6.5395 & 7.0073 & 13.7524 & \underline{\textbf{54.3200}} & 2.4067 & 0.8804 & 1.2224 & \underline{0.7538} & 1.0625 \\
DQN & \underline{-0.7291} & 1.8888 & \underline{1.4548} & 3.5727 & 13.2428 & 2.0431 & 1.3333 & 1.9119 & 11.6727 & 16.9483 \\
M-FOS & -1.0666 & \underline{1.7833} & 2.1642 & \underline{3.0970} & 15.9722 & \underline{3.2033} & \underline{1.1563} & \underline{1.9719} & 2.4573 & \underline{4.0411} \\
Abl. M-FOS & -5.0886 & 17.6561 & 2.2476 & 5.4770 & 27.9635 & 5.3393 & 2.0600 & 3.5372 & 3.9810 & 5.9086 \\
\bottomrule
\end{tabular}
\end{subtable}

\begin{subtable}{\textwidth}
\caption{Simulation Data, 1M Samples, Solo Average Performance}
\label{tab:indSim}
\centering
\small
\begin{tabular}{lcccccccccc}
\toprule
Agent & $\bar{R}\uparrow$ & $\sigma_R\downarrow$ & $\bar{C}\downarrow$ & $\sigma_C\downarrow$ & $\overline{\mathrm{BW}}\uparrow$ & $\sigma_{BW}\downarrow$ & $\Delta BW\downarrow$ & $\sigma_{\Delta BW}\downarrow$ & $\Delta CF\downarrow$ & $\sigma_{\Delta CF}\downarrow$ \\
\midrule
Random Static & -0.9946 & 2.7879 & 2.3437 & 5.6647 & 24.3555 & \textbf{0} & \textbf{0} & \textbf{0} & \textbf{0} & \textbf{0} \\
SAA & \textbf{-0.1093} & 1.7045 & \textbf{0.6803} & 3.1050 & \textbf{69.7126} & 23.7323 & 19.0551 & 18.5211 & 12.8631 & 14.0145 \\
\midrule
PPO & -1.5682 & 3.7726 & 3.6773 & 7.9365 & \underline{34.6500} & \underline{1.0123} & 0.5678 & \underline{0.7778} & \underline{0.4736} & \underline{0.5546} \\
DQN & -0.5772 & 1.6253 & 1.0921 & 3.0228 & 12.3288 & 2.5295 & 2.0601 & 2.3923 & 19.1770 & 20.0172 \\
M-FOS & \underline{-0.3806} & \underline{\textbf{1.5390}} & \underline{0.8565} & \underline{\textbf{2.6819}} & 14.6190 & 3.3200 & 2.0608 & 2.1255 & 1.1812 & 1.3807 \\
Abl. M-FOS & -5.5613 & 17.5247 & 2.4772 & 5.4772 & 25.9548 & 1.0734 & \underline{0.3507} & 0.9963 & 1.6558 & 2.5561 \\
\bottomrule
\end{tabular}
\end{subtable}

\begin{subtable}{\textwidth}
\caption{2.4--2.5GHz Memorial Union Data, 4.1M Samples, Solo Average Performance}
\label{tab:ind245union}
\centering
\small
\begin{tabular}{lcccccccccc}
\toprule
Agent & $\bar{R}\uparrow$ & $\sigma_R\downarrow$ & $\bar{C}\downarrow$ & $\sigma_C\downarrow$ & $\overline{\mathrm{BW}}\uparrow$ & $\sigma_{BW}\downarrow$ & $\Delta BW\downarrow$ & $\sigma_{\Delta BW}\downarrow$ & $\Delta CF\downarrow$ & $\sigma_{\Delta CF}\downarrow$ \\
\midrule
Random Start & -0.6825 & 1.8853 & 1.5495 & 3.6076 & 15.1693 & \textbf{0} & \textbf{0} & \textbf{0} & \textbf{0} & \textbf{0} \\
SAA & \textbf{0.0463} & 1.3071 & \textbf{0.2676} & \textbf{1.3717} & \textbf{58.6795} & 22.7560 & 18.3399 & 17.6984 & 13.6188 & 14.4751 \\
\midrule
PPO & -0.9673 & \underline{\textbf{1.2108}} & 2.0805 & \underline{1.5307} & 19.0346 & 3.0575 & \underline{1.5462} & \underline{1.5838} & \underline{1.6535} & \underline{1.6534} \\
DQN & -0.5469 & 1.4494 & \underline{1.0047} & 2.5264 & 13.1032 & \underline{2.4418} & 1.8237 & 2.2372 & 20.5137 & 22.8263 \\
M-FOS & \underline{-0.4944} & 1.8335 & 1.0400 & 3.3898 & 15.5820 & 6.2020 & 2.6591 & 4.6239 & 3.2751 & 4.9135 \\
Abl. M-FOS & -6.5595 & 27.0167 & 1.8829 & 4.9427 & \underline{30.6390} & 6.4058 & 2.7083 & 3.7575 & 2.9423 & 3.9461 \\
\bottomrule
\end{tabular}
\end{subtable}

\begin{subtable}{\textwidth}
\caption{2.59--2.69GHz Memorial Union Data, 2.4M Samples, Solo Average Performance}
\label{tab:ind264union}
\centering
\small
\begin{tabular}{lcccccccccc}
\toprule
Agent & $\bar{R}\uparrow$ & $\sigma_R\downarrow$ & $\bar{C}\downarrow$ & $\sigma_C\downarrow$ & $\overline{\mathrm{BW}}\uparrow$ & $\sigma_{BW}\downarrow$ & $\Delta BW\downarrow$ & $\sigma_{\Delta BW}\downarrow$ & $\Delta CF\downarrow$ & $\sigma_{\Delta CF}\downarrow$ \\
\midrule
Random Start & -0.6828 & 1.9065 & 1.5504 & 3.6147 & 15.1693 & \textbf{0} & \textbf{0} & \textbf{0} & \textbf{0} & \textbf{0} \\
SAA & \textbf{0.0462} & 1.3027 & 0.2677 & 1.3701 & \textbf{58.6690} & 22.6520 & 18.3705 & 17.6344 & 13.7221 & 14.4765 \\
\midrule
PPO & \underline{-0.0677} & \underline{\textbf{0.9092}} & \underline{\textbf{0.2205}} & \underline{\textbf{0.7542}} & 10.3898 & 0.4971 & 0.1930 & \underline{0.2421} & \underline{0.5249} & \underline{0.5453} \\
DQN & -0.6661 & 1.6501 & 1.2571 & 3.0675 & 12.4909 & 2.5351 & 1.9726 & 2.3763 & 20.5427 & 22.7386 \\
M-FOS & -0.3221 & 1.2666 & 0.6388 & 2.0137 & 9.9905 & \underline{0.2798} & \underline{0.0501} & 0.2997 & 4.0354 & 5.8188 \\
Abl. M-FOS & -6.6563 & 25.8445 & 1.9381 & 5.0380 & \underline{31.4457} & 5.2342 & 2.1702 & 3.0761 & 2.9208 & 4.0044 \\
\bottomrule
\end{tabular}
\end{subtable}
\end{table*}
Although M-FOS does not achieve the highest reward in every dataset, it consistently ranks among the best-performing learning agents while exhibiting comparatively stable behavior across environments. In contrast, PPO and DQN achieve the highest reward only under specific spectrum conditions. The ablated M-FOS performs considerably worse than the full architecture in every environment, suggesting that the proposed meta-learning mechanism contributes significantly to policy stability and overall performance.

The action statistics further distinguish the learning approaches. PPO generally learns to occupy bandwidths comparable to those selected by SAA while maintaining very small changes in bandwidth and center frequency between consecutive actions. However, this conservative behavior is accompanied by relatively poor collision avoidance, resulting in substantially lower rewards than either DQN or M-FOS. DQN exhibits the opposite tendency, selecting much narrower bandwidths while incurring relatively large center-frequency changes. M-FOS instead learns a balanced operating strategy, typically selecting narrower bandwidths than PPO but exhibiting consistently smaller changes in both bandwidth and center frequency than DQN. This balance enables M-FOS to maintain competitive collision rates while simultaneously reducing waveform distortion metrics.

The consistency of these trends across both measured over-the-air datasets and the simulation environment suggests that the proposed architecture generalizes well across different spectrum occupancy characteristics rather than overfitting to a particular dataset.

\subsection{Multi-Agent Results}

To further compare the performance of each cognitive agent in a multi-agent environment, experiments were conducted for the 2.4--2.5~GHz, 2.59-2.69~GHz, and simulation environments with 3 of each agent acting simultaneously. Preliminary results indicate that M-FOS achieves superior average reward in multi-agent environments with both measured over-the-air and simulated data against all compared agents. The results are shown in Tables~\ref{tab:245MultiAgent} --~\ref{tab:u264MultiAgent}. These tables also include results from running PPO, DQN, and M-FOS simultaneously to be able to compare average results when the same algorithm is used against results where three different algorithms interact with each other. The best values in each category are highlighted in green with the reward values generally indicating an agent's ability to balance competing objectives. The experiments in which each agent is evaluated alongside two additional agents of the same type indicate that M-FOS achieves the highest average reward in most scenarios (except in Table~\ref{tab:simMultiAgent} where the ablated version slightly outperforms it). In the two union dataset experiments (Tables~\ref{tab:u245MultiAgent} and~\ref{tab:u264MultiAgent}), the mixing of agent types actually achieved the best average reward values (by small margins over the M-FOS triplets).
 
\begin{figure}[htbp]
    \centering
    \subfloat[Multi-Agent Average Rewards
    \label{maMeanReward}]{
        \includegraphics[width=\linewidth]{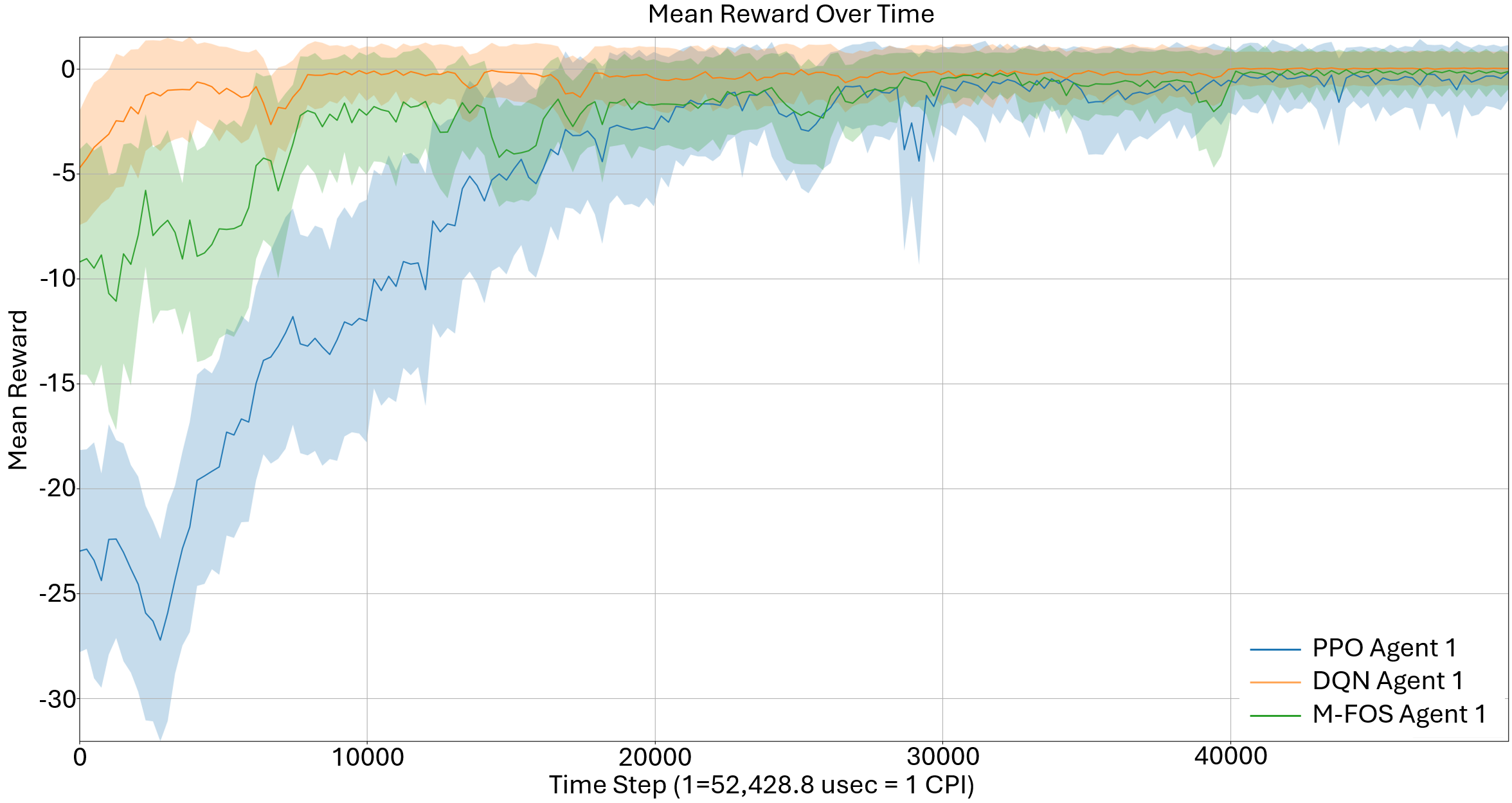}
    }
    \hfill
    \subfloat[Solo Agent Average Rewards
    \label{soloMeanReward}]{
        \includegraphics[width=\linewidth]{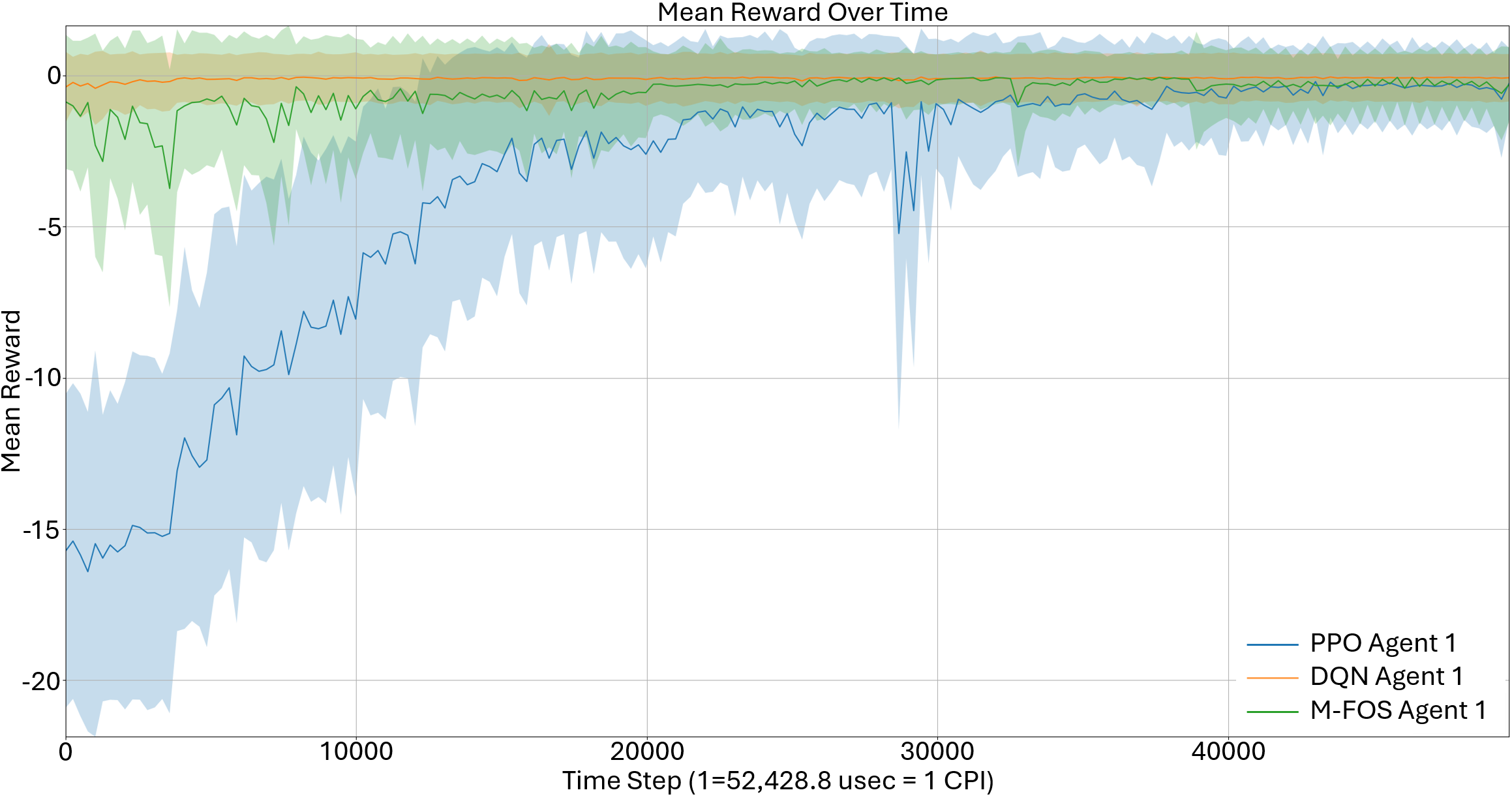}
    }
    \caption{Comparison of average reward performance.}
    \label{fig:AveRewardCompared}
\end{figure}

 The multi-agent plots in Fig.~\ref{fig:AveRewardCompared} on its own does not demonstrate superiority of any agent type as the values and speeds of convergence can vary significantly based on random starting points. For example, if the starting time index of DQN is prior to that of the other agents, then only the other two agents will generally receive penalties when intruding on the DQN agent's transmission space, since bin ownership is based on first-come, first-served. In these multi-agent settings, it is likely best to consider the average performance and cooperative results rather than evaluating agents competitively. However, the solo agent plots from the same figures generally demonstrate similar convergence speeds to the multi-agent plots. These results suggest somewhat slower convergence for M-FOS compared to DQN, and much slower convergence speeds for the PPO agent; however, if the PPO agent had initialized with a smaller bandwidth value, it may have been able to converge much faster, so the graphical results from single experiments are limited in what they demonstrate. Convergence results from the reward plots in \cite{flandermeyer2024} show faster and smoother convergence for PPO over DQN, but, given slightly different implementations of these algorithms, action spaces, and datasets, differing results are to be expected. More data and fine-tuning of the models is necessary to draw conclusions about convergence speeds.

\begin{table*}[htbp]
\centering
\caption{Performance comparison of all methods. Each value represents the average  of three values per agent type.}
\label{tab:multiResults}
\begin{subtable}{\textwidth}
\caption{ARRC Data 2.4--2.5~GHz, Multi-Agent Triplet Average Performance}
\label{tab:245MultiAgent}
\centering
\small
\begin{tabular}{lcccccccccc}
\toprule
Agent & $\bar{R}\uparrow$ & $\sigma_R\downarrow$ & $\bar{C}\downarrow$ & $\sigma_C\downarrow$ & $\overline{\mathrm{BW}}\uparrow$ & $\sigma_{BW}\downarrow$ & $\Delta BW\downarrow$ & $\sigma_{\Delta BW}\downarrow$ & $\Delta CF\downarrow$ & $\sigma_{\Delta CF}\downarrow$ \\
\midrule
SAA & -20.2359 & 7.0489 & 40.7458 & 13.806 & \textbf{61.1295} & 20.4686 & 17.4796 & 15.5938 & 15.6053 & 15.7289 \\
PPO  & -0.336 & 0.9297 & 13.8559 & 4.7051 & 42.5781 & \textbf{0.0003} & \textbf{0.0451} & \textbf{0.1055} & \textbf{0.0208} & \textbf{0.0484} \\
DQN  & -0.0792 & 0.4050 & 3.1832 & 4.1661 & 12.3156 & 2.6327 & 2.4554 & 2.4503 & 24.4876 & 23.1062 \\
M-FOS & \textbf{-0.0493} & \textbf{0.2612} & 2.0186 & 1.1672 & 10.1811 & 1.5085 & 0.1203 & 0.7876 & 0.0906 & 0.3214 \\
Abl. M-FOS & -0.1729 & 0.5537 & 2.3313 & 1.9785 & 21.9816 & 0.2808 & 0.1553 & 0.1673 & 0.5758 & 0.7293 \\
PPO/DQN/M-FOS & -0.5574 & 0.9597 & \textbf{1.1996} & \textbf{1.0878} & 12.8298 & 1.4777 & 0.9771 & 1.0866 & 2.8808 & 5.6228 \\
\bottomrule
\end{tabular}
\end{subtable}

\begin{subtable}{\textwidth}
\caption{ARRC Data 2.59--2.69~GHz, Multi-Agent Triplet Average Performance}
\label{tab:264MultiAgent}
\centering
\small
\begin{tabular}{lcccccccccc}
\toprule
Agent & $\bar{R}\uparrow$ & $\sigma_R\downarrow$ & $\bar{C}\downarrow$ & $\sigma_C\downarrow$ & $\overline{\mathrm{BW}}\uparrow$ & $\sigma_{BW}\downarrow$ & $\Delta BW\downarrow$ & $\sigma_{\Delta BW}\downarrow$ & $\Delta CF\downarrow$ & $\sigma_{\Delta CF}\downarrow$ \\
\midrule
SAA & -15.7775 & 11.6817 & 32.1563 & 24.0067 & \textbf{54.044} & 23.0169 & 18.4608 & 19.6702 & 16.6488 & 16.9795 \\
PPO  & -0.2485 & 0.9009 & 10.3650 & 6.7079 & 32.9889 & 0.3846 & 0.2903 & 0.4003 & 0.2781 & 0.3982 \\
DQN  & -0.1084 & 0.4890 & 4.4234 & 4.8713 & 12.8432 & 2.2223 & 1.8124 & 2.1495 & 18.7139 & 21.4243 \\
M-FOS & \textbf{-0.0269} & \textbf{0.2650} & \textbf{1.1571} & \textbf{2.3945} & 9.9609 & \textbf{0.0000} & \textbf{0.0264} & \textbf{0.1109} & \textbf{0.2061} & \textbf{0.3478} \\
Abl. M-FOS & -0.2604 & 0.8833 & 2.0839 & 3.1603 & 13.4941 & 2.2713 & 0.8757 & 1.3984 & 0.7608 & 1.2358 \\
PPO/DQN/M-FOS & -1.1460 & 2.0633 & 2.435 & 3.4916 & 17.9334 & 0.8549 & 0.6772 & 0.7746 & 2.0397 & 5.0005 \\
\bottomrule
\end{tabular}
\end{subtable}

\begin{subtable}{\textwidth}
\caption{Simulation Data, Multi-Agent Triplet Average Performance, 1M Samples}
\label{tab:simMultiAgent}
\centering
\small
\begin{tabular}{lcccccccccc}
\toprule
Agent & $\bar{R}\uparrow$ & $\sigma_R\downarrow$ & $\bar{C}\downarrow$ & $\sigma_C\downarrow$ & $\overline{\mathrm{BW}}\uparrow$ & $\sigma_{BW}\downarrow$ & $\Delta BW\downarrow$ & $\sigma_{\Delta BW}\downarrow$ & $\Delta CF\downarrow$ & $\sigma_{\Delta CF}\downarrow$ \\
\midrule
SAA  & -1.1068 & 3.3842 & 45.1176 & 25.5904 & \textbf{71.7509} & 21.7112 & 18.1899 & 17.7005 & 13.0895 & 13.9095 \\
PPO  & -0.2772 & 0.8141 & 11.4324 & 2.2243 & 36.7165 & 0.0685 & 0.0675 & 0.1356 & 0.1528 & 0.2684 \\
DQN  & -0.0726 & 0.3574 & 2.9586 & 2.9050 & 13.8001 & 1.4421 & 1.0234 & 1.5046 & 14.9609 & 15.4186 \\
M-FOS & -0.0070 & \textbf{0.2130} & 0.3710 & 1.2093 & 9.9609 & \textbf{0} & 0.0270 & 0.1122 & \textbf{0.0657} & \textbf{0.1328} \\
Abl. M-FOS & \textbf{0.0113} & 0.3853 & \textbf{0.1097} & \textbf{0.5603} & 9.9609 & \textbf{0} & \textbf{0.0141} & \textbf{0.0641} & 0.3025 & 0.3898 \\
PPO/DQN/M-FOS & -0.1324 & 0.8691 & 0.3168 & 0.8191 & 11.173 & 1.6434 & 0.7273 & 0.9126 & 4.125 & 8.5537 \\
\bottomrule
\end{tabular}
\end{subtable}

\begin{subtable}{\textwidth}
\caption{Union Data 2.4--2.5~GHz, Multi-Agent Triplet Average Performance, 4.1M Samples}
\label{tab:u245MultiAgent}
\centering
\small
\begin{tabular}{lcccccccccc}
\toprule
Agent & $\bar{R}\uparrow$ & $\sigma_R\downarrow$ & $\bar{C}\downarrow$ & $\sigma_C\downarrow$ & $\overline{\mathrm{BW}}\uparrow$ & $\sigma_{BW}\downarrow$ & $\Delta BW\downarrow$ & $\sigma_{\Delta BW}\downarrow$ & $\Delta CF\downarrow$ & $\sigma_{\Delta CF}\downarrow$ \\
\midrule
SAA & -7.3495 & 5.8718 & 14.925 & 11.9945 & \textbf{27.4071} & 9.1130 & 7.7232 & 7.8254 & 20.1869 & 21.1601 \\
PPO & -0.7348 & 1.1527 & 1.5587 & 1.7772 & 10.9169 & 0.3552 & 0.1619 & 0.1597 & 0.4206 & 0.3905 \\
DQN & -2.0741 & 2.0809 & 4.1773 & 4.1539 & 12.4331 & 2.5420 & 2.5775 & 2.3296 & 28.2459 & 22.1603 \\
M-FOS & -0.4879 & 1.0440 & 1.0642 & 1.3419 & 9.9609 & \textbf{0} & \textbf{0.0095} & \textbf{0.0191} & \textbf{0.2079} & \textbf{0.2152} \\
Abl. M-FOS & -0.6506 & 2.8926 & \textbf{0.7156} & \textbf{1.0762} & 9.9634 & 0.1201 & 0.0138 & 0.0603 & 1.8114 & 1.8260 \\
PPO/DQN/M-FOS & \textbf{-0.4561} & \textbf{1.0047} & 0.9838 & 1.3448 & 10.3242 & 0.6903 & 0.6340 & 0.6927 & 1.9349 & 4.3633 \\
\bottomrule
\end{tabular}
\end{subtable}

\begin{subtable}{\textwidth}
\caption{Union Data 2.59--2.69~GHz, Multi-Agent Triplet Average Performance, 2.4M Samples}
\label{tab:u264MultiAgent}
\centering
\small
\begin{tabular}{lcccccccccc}
\toprule
Agent & $\bar{R}\uparrow$ & $\sigma_R\downarrow$ & $\bar{C}\downarrow$ & $\sigma_C\downarrow$ & $\overline{\mathrm{BW}}\uparrow$ & $\sigma_{BW}\downarrow$ & $\Delta BW\downarrow$ & $\sigma_{\Delta BW}\downarrow$ & $\Delta CF\downarrow$ & $\sigma_{\Delta CF}\downarrow$ \\
\midrule
SAA & -17.7055 & 11.0699 & 35.9245 & 22.5762 & \textbf{58.7219} & 22.7352 & 18.3219 & 17.6243 & 13.6994 & 14.5431 \\
PPO & -0.9534 & 2.6429 & 2.3108 & 5.2800 & 26.5951 & 0.0164 & 0.0103 & 0.0191 & \textbf{0.0102} & \textbf{0.0182} \\
DQN & -1.2724 & 2.0274 & 2.6109 & 4.0137 & 11.9580 & 2.4038 & 1.8975 & 2.2835 & 18.5236 & 20.1513 \\
M-FOS & -0.3257 & 1.2947 & 0.7281 & 2.1189 & 9.9609 & \textbf{0} & \textbf{0.0044} & \textbf{0.0152} & 0.3643 & 0.5175 \\
Abl. M-FOS & -3.0242 & 6.2672 & 3.6979 & 3.4014 & 18.2530 & 1.5246 & 0.7175 & 0.8386 & 0.6161 & 0.8095 \\
PPO/DQN/M-FOS & \textbf{-0.2846} & \textbf{1.2172} & \textbf{0.606} &\textbf {1.7917} & 10.5353 & 0.8230 & 0.7615 & 0.7711 & 3.5157 & 6.1861 \\
\bottomrule
\end{tabular}
\end{subtable}
\end{table*}

\begin{figure}[htbp]
\centering
\includegraphics[width=1\linewidth]{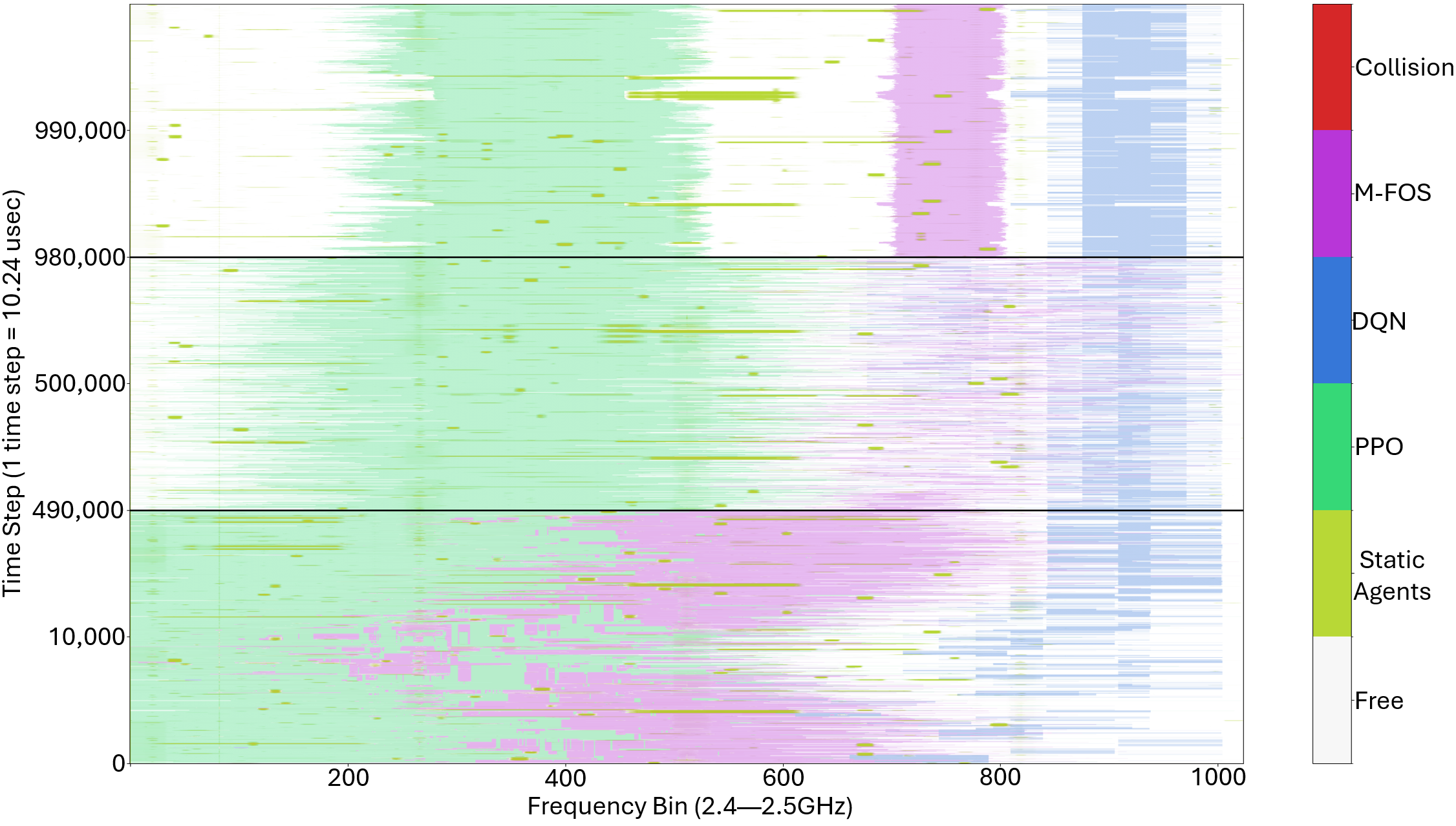}
\caption{Multi-Agent Spectrum Occupancy between PPO, M-FOS, and DQN. First, middle, and last 20,000 Samples, from bottom to top respectively.}
\label{fig:maSpectrumOccupancy}
\end{figure}

Agent behavior in a multi-agent experiment with PPO, DQN, and M-FOS is visually demonstrated in Fig.~\ref{fig:maSpectrumOccupancy}. The PPO agent transmits the widest bandwidth and shifts only in response to relatively long-lasting static transmissions. The M-FOS agent chooses a smaller bandwidth, slightly more variable with its willingness to adapt its transmission. The DQN agent occupies comparatively less congested spectral regions and shows the highest amount of variability in its transmissions. Note that all agents learned to successfully avoid each others' transmissions.

Additional plots illustrate convergence behavior in both solo and multi-agent environments for the same three agents in Fig.~\ref{fig:AveRewardCompared}. The DQN agent (orange) starts with the highest starting rewards, and quickly converges after roughly 8,000 CPIs. The M-FOS agent has intermediate initial reward values and occasionally has temporary large dips during the training period reflecting exploratory behavior induced by genome switching; it converges at about the same speed as the DQN agent. The PPO agent has the worst starting returns (likely from exploring very large bandwidths that result in large collisions), and slowly converges after approximately 25,000--30,000 CPIs.

\subsection{Transfer Learning}
\begin{figure}[htbp]
    \centering
    \subfloat[Average Reward in Multi-Agent Environment Over Time
    \label{fig:245MultiMeanReward}]{
        \includegraphics[width=.95\linewidth]{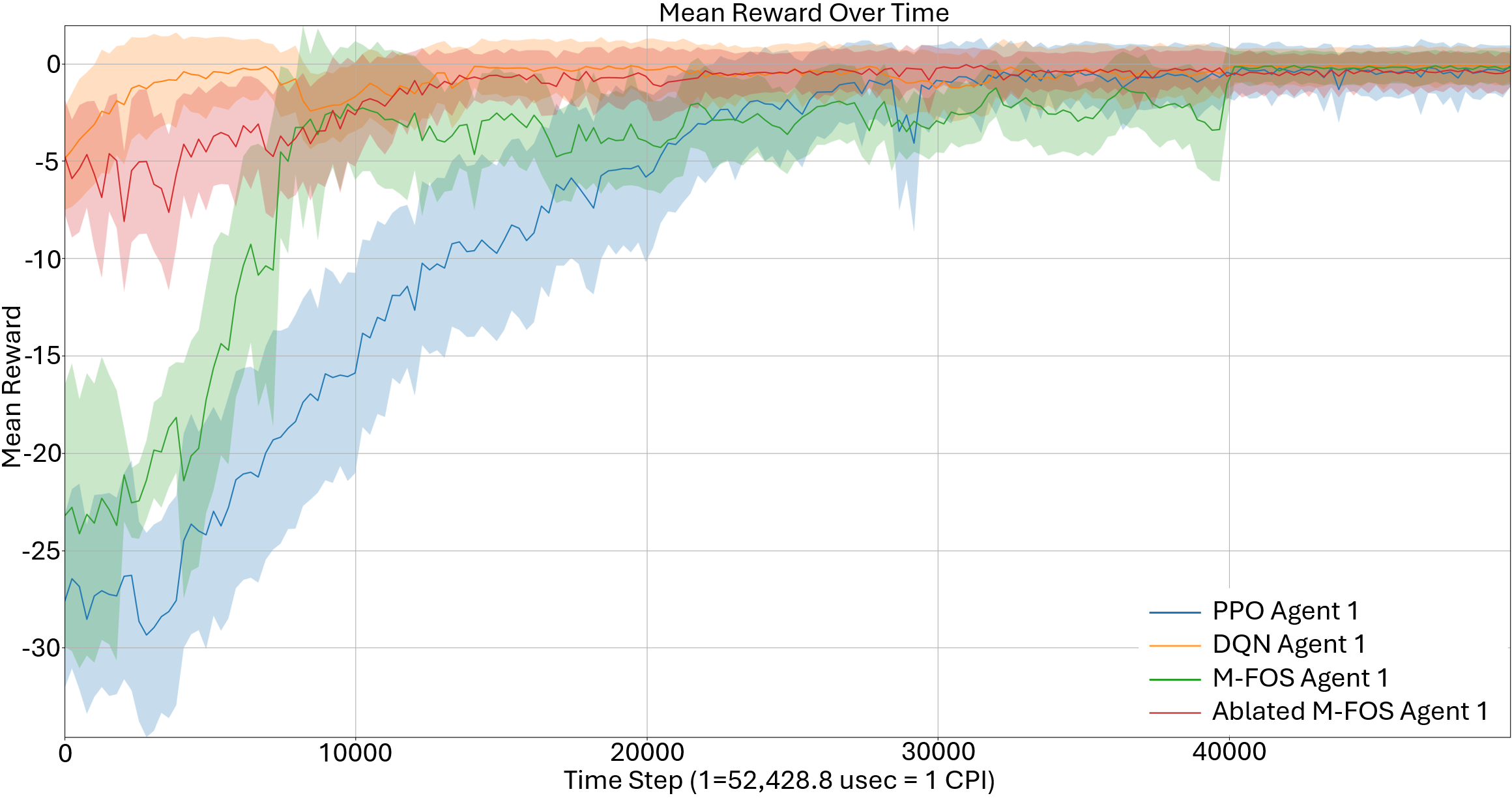}
    }
    \hfill
    \subfloat[Average Reward in Multi-Agent Environment Over Time Using Pretrained Models
    \label{fig:245MultiPretrainedMeanReward}]{
        \includegraphics[width=.95\linewidth]{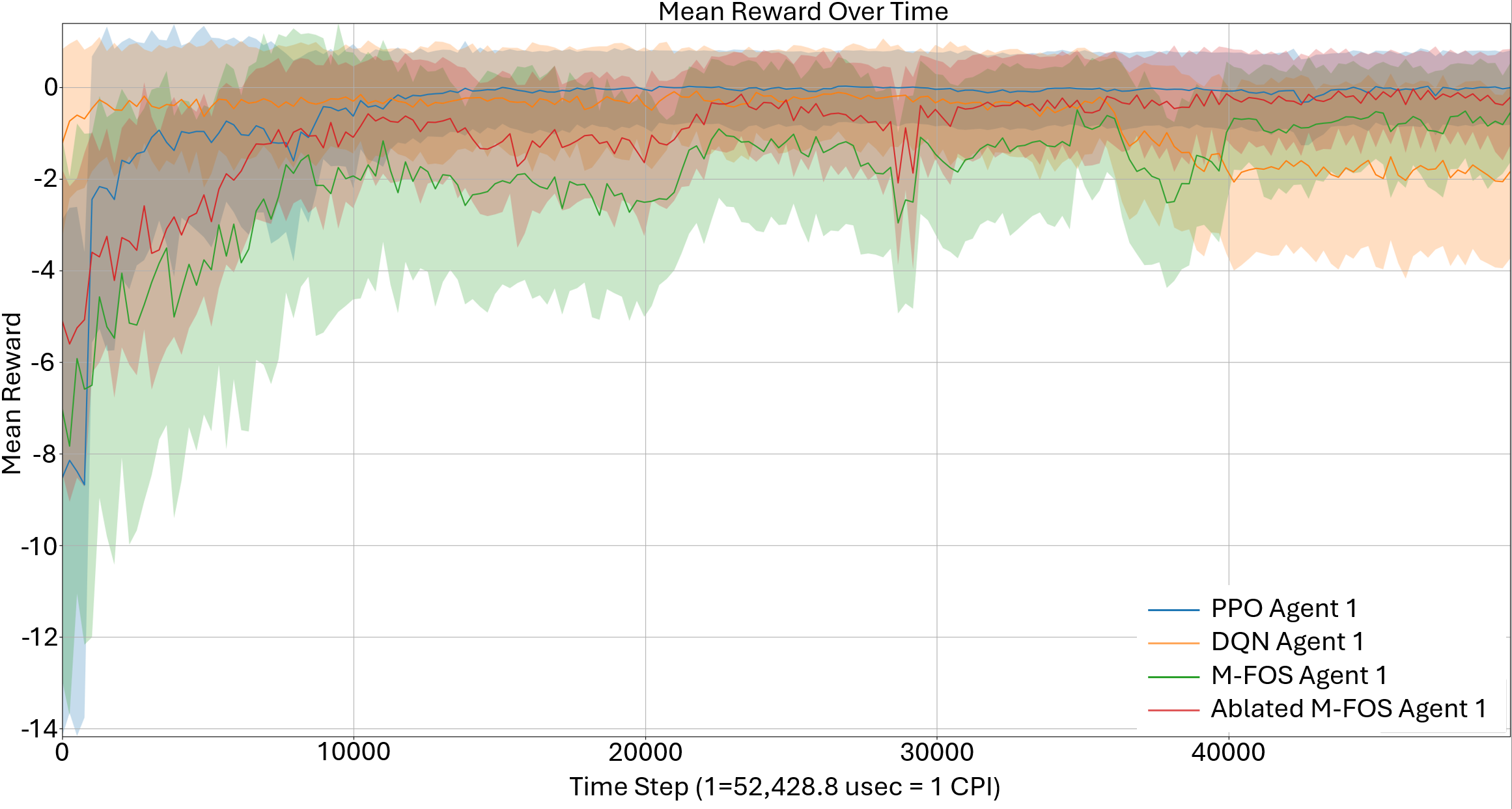}
    }
    \caption{Pretrained model comparison of average reward performance in the multi-agent environment using ARRC 2.4--2.5~GHz dataset.}
    \label{fig:245MultiRewardComparison}
\end{figure}

\begin{figure}[htbp]
    \centering
    \subfloat[Average Reward in Multi-Agent Environment Over Time
    \label{fig:u245MultiMeanReward}]{
        \includegraphics[width=\linewidth]{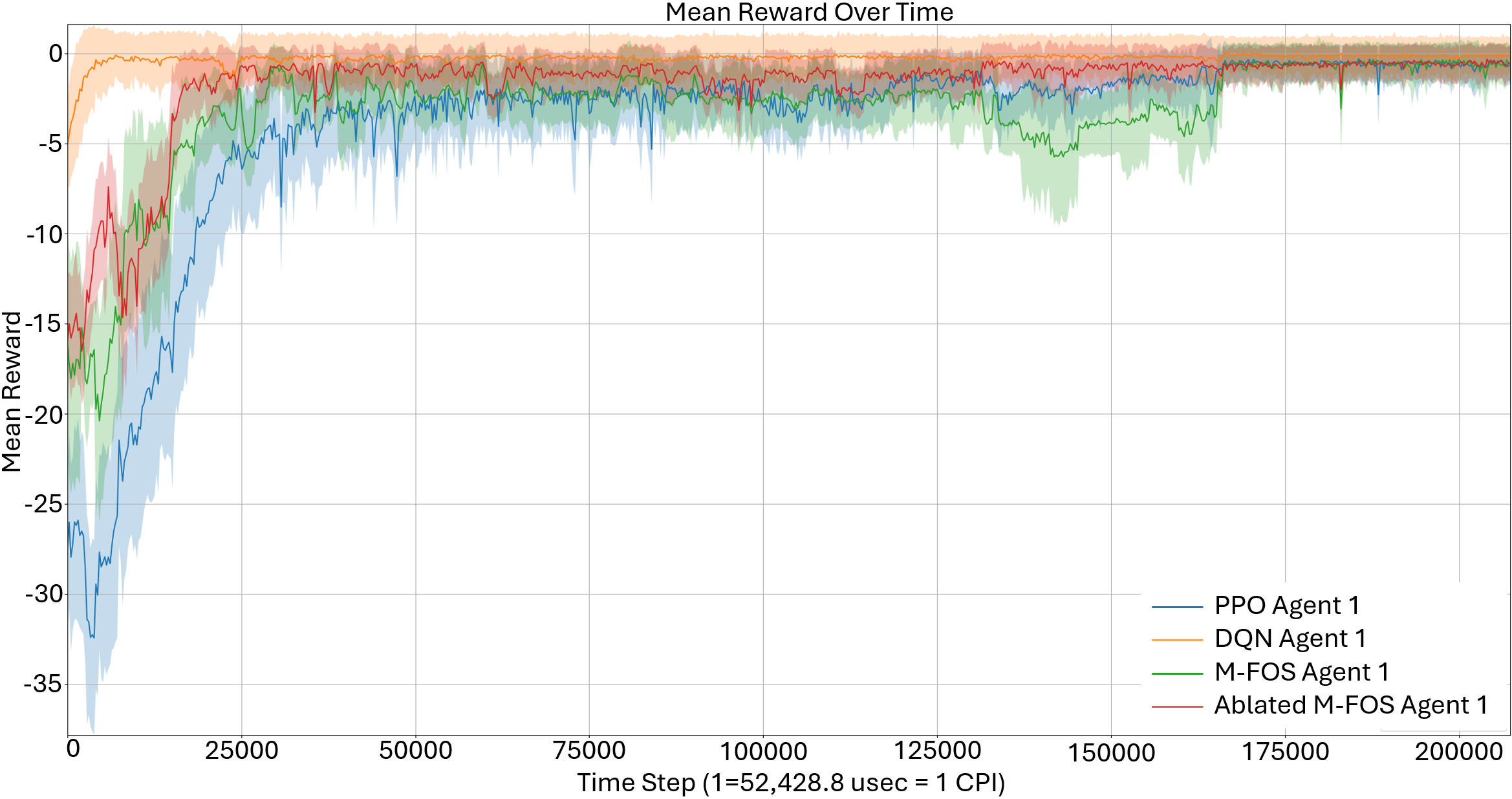}
    }
    \hfill
    \subfloat[Average Reward in Multi-Agent Environment Over Time Using Pretrained Models
    \label{fig:u245MultiPretrainedMeanReward}]{
        \includegraphics[width=\linewidth]{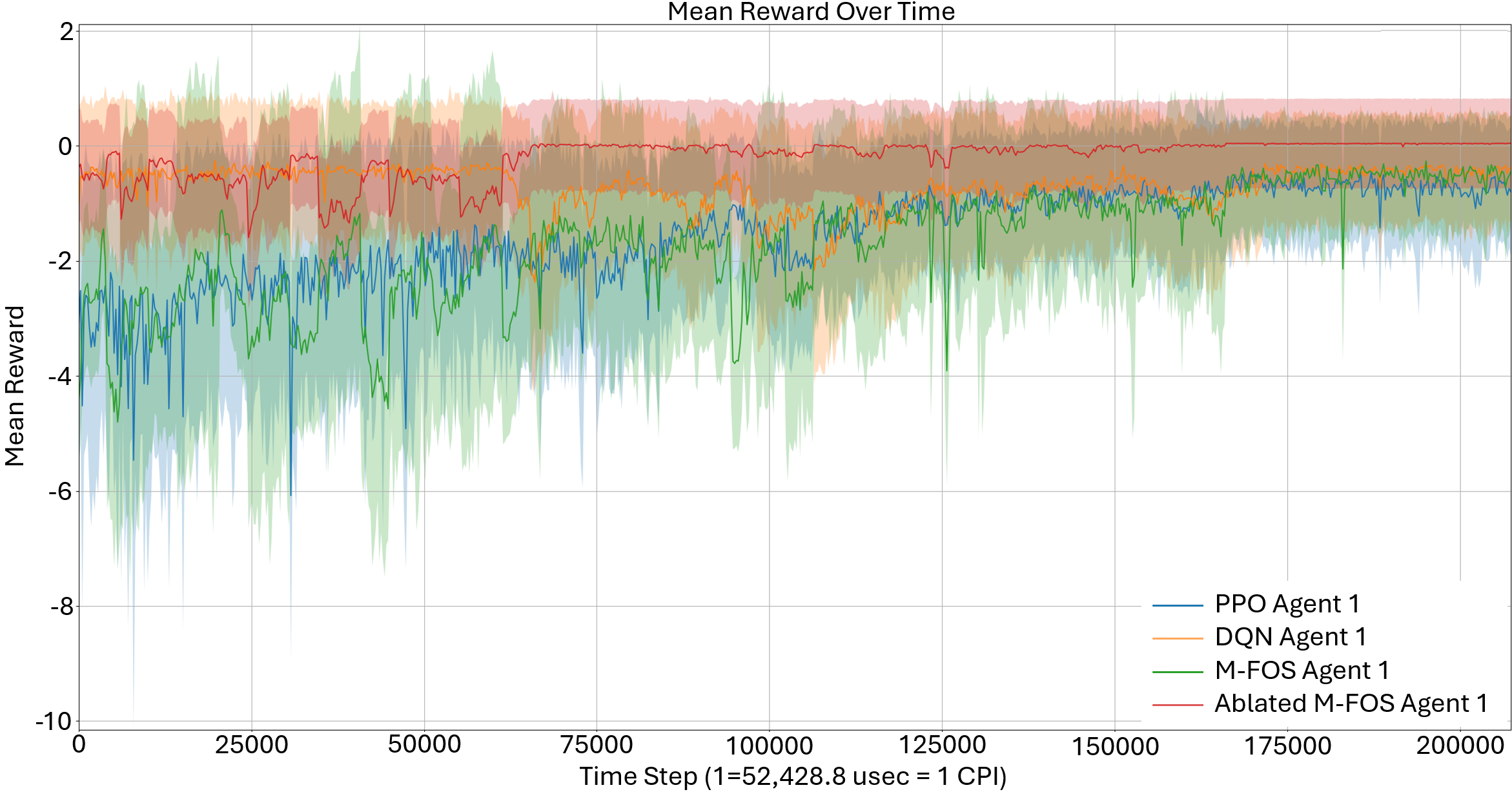}
    }
    \caption{Pretrained model comparison of average reward performance in the multi-agent environment using Union 2.4--2.5~GHz dataset.}
    \label{fig:u245multiRewardComparison}
\end{figure}

Experiments were also conducted by training models in solo environments then comparing multi-agent experiments using the pretrained models against completely untrained agents. Note Figs.~\ref{fig:245MultiRewardComparison} and~\ref{fig:u245multiRewardComparison} in which the convergence times are drastically reduced, and the average reward begins in a much better place. PPO for example improves from an average starting reward of -28 to just -8, and it quickly jumps to -1 after about 25,000 CPIs. PPO's convergence time decreases from approximately 300,000 to 110,000 CPIs. The other agents show similar improvements in initial average rewards and in convergence speeds. This demonstrates that there is enough commonality between multi-agent and single-agent environments for a policy to quickly adapt from one to the other. It also suggests that good initialization of the policy weights has a significant impact on short-term performance.

\subsection{Range-Doppler Responses}

\begin{figure}[htbp]
    \centering
    \subfloat[SAA Range-Doppler Plot
    \label{saaRngVel}]{
        \includegraphics[width=.45\linewidth]{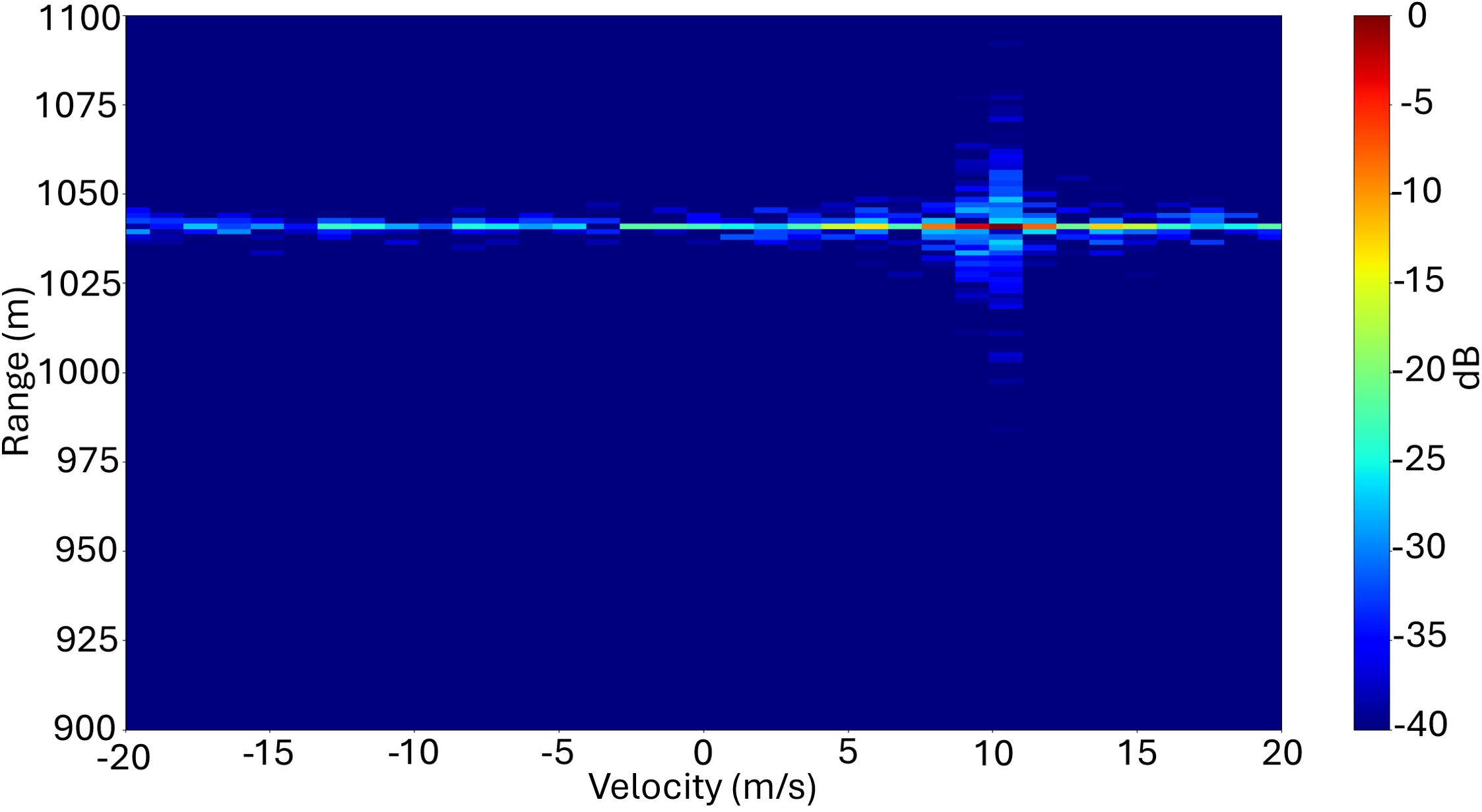}
    }
    \hfill
    \subfloat[PPO Range-Doppler Plot
    \label{ppoRngVel}]{
        \includegraphics[width=.45\linewidth]{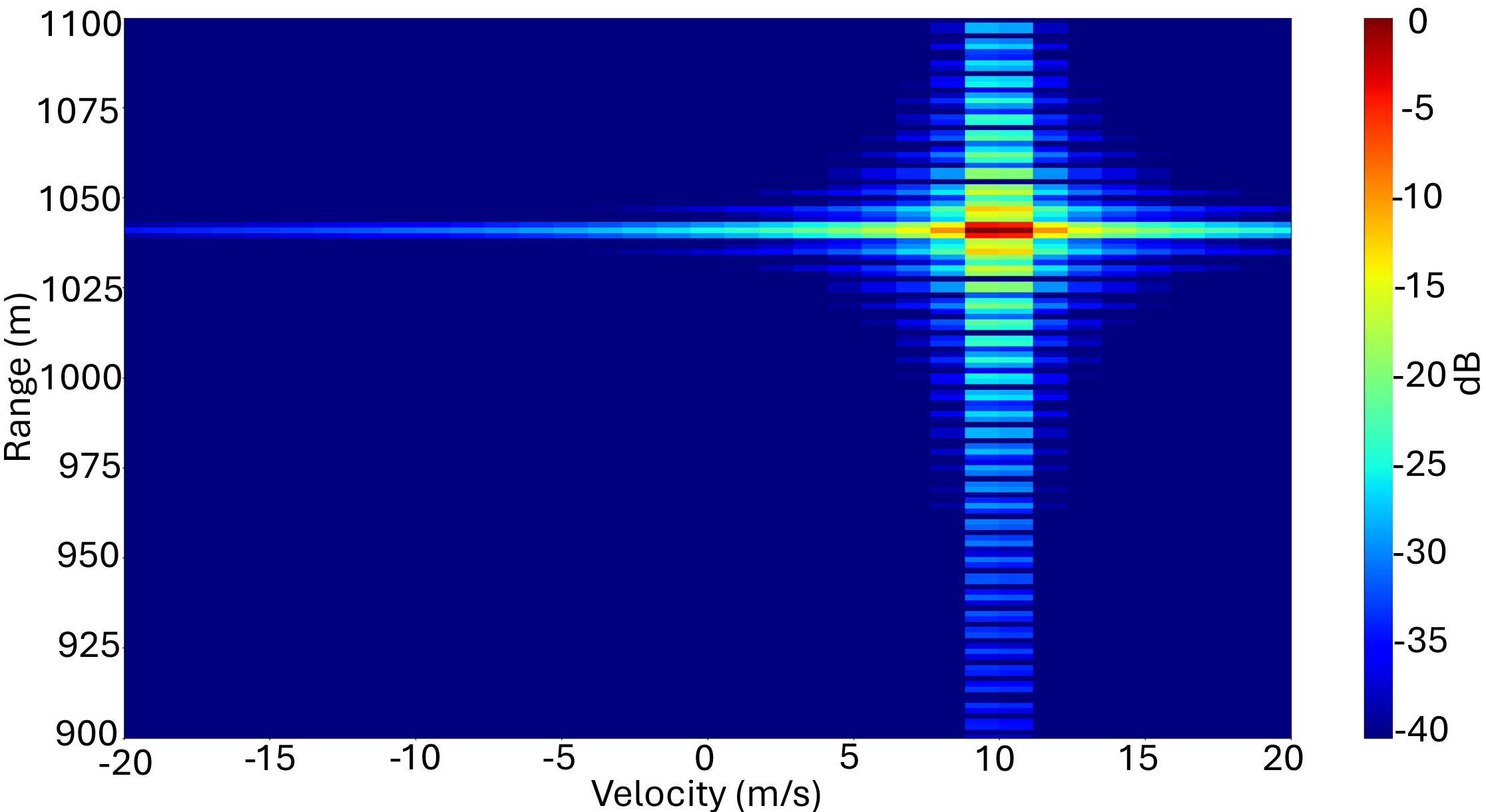}
    }
    \hfill
    \subfloat[DQN Range-Doppler Plot
    \label{dqnRngVel}]{
        \includegraphics[width=.45\linewidth]{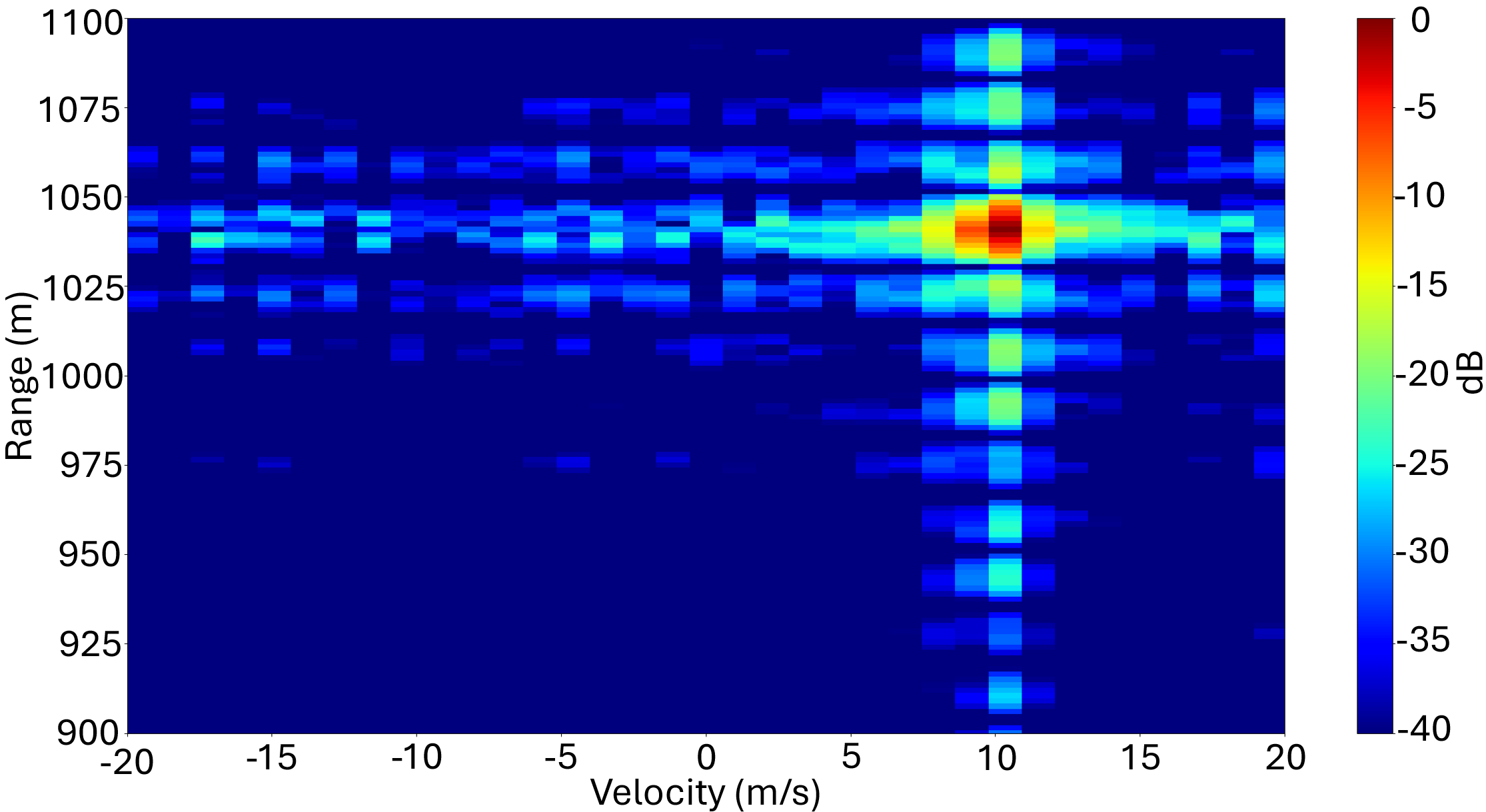}
    }
    \hfill
    \subfloat[M-FOS Range-Doppler Plot
    \label{mfosRngVel}]{
        \includegraphics[width=.45\linewidth]{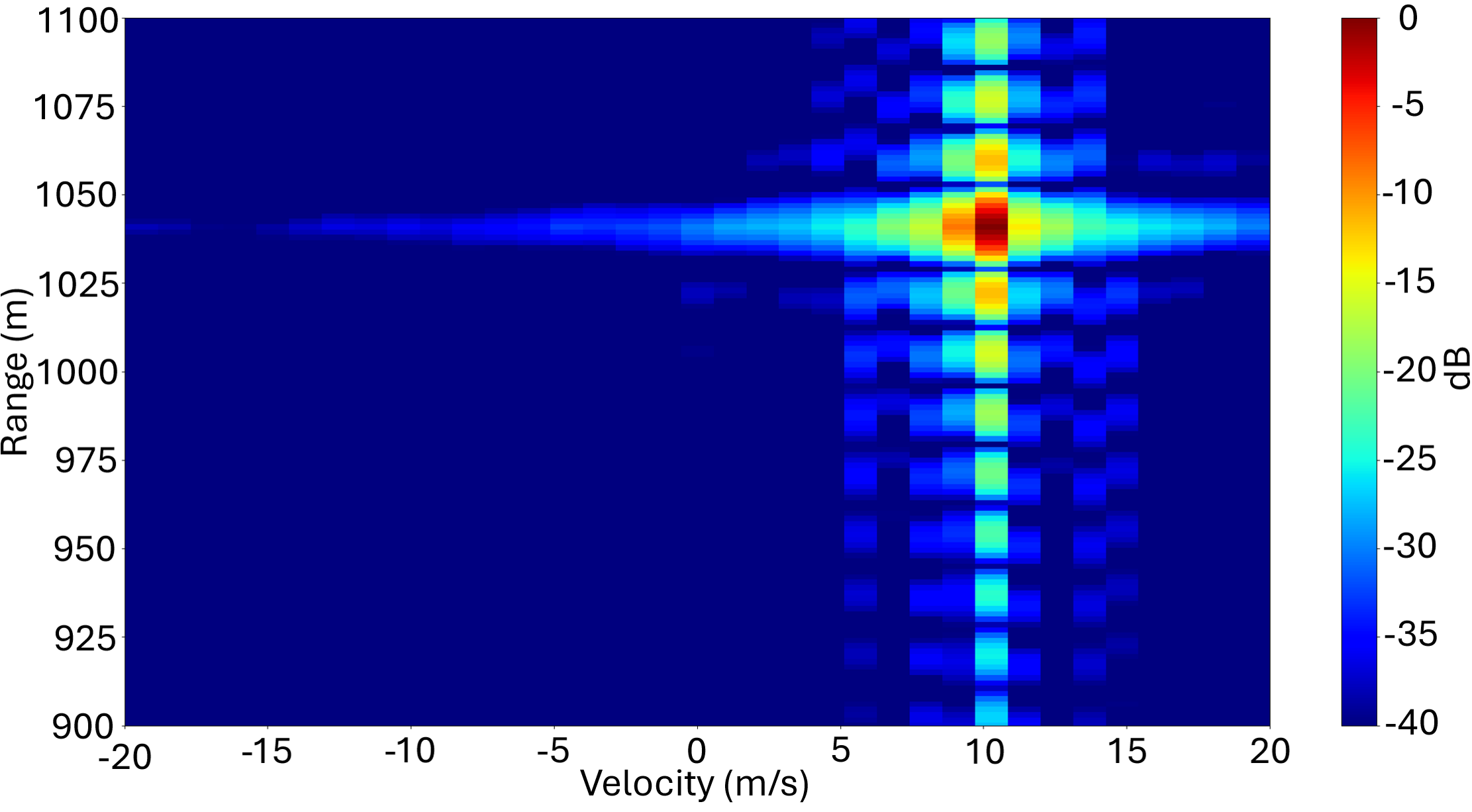}
    }
    \caption{Range Doppler Plots}
    \label{fig:rngVelPlots}
\end{figure}

In Table~\ref{tab:distortionStats}, an experiment with five different agents with various distortion statistics is used to create simulated Range-Doppler responses in Fig.~\ref{fig:rngVelPlots}. PPO exhibits low, consistent distortion metrics as evidenced by the clarity of the response in Fig.~\ref{ppoRngVel}, and contrast this with SAA's plot in Fig.~\ref{saaRngVel}. The CPI chosen for these plots was the first full CPI from the evaluation period of the experiment, and the evaluation period was reduced from 20\% of the dataset to just 1\% (approximately 2 CPIs) in order to best link the averaged statistics from the evaluation period with the range-Doppler plot from a single CPI.

\begin{table}[htbp]
\caption{Multi-Agent Distortion Statistics 2.4--2.5~GHz ARRC Data}
\label{tab:distortionStats}
\centering
\small
\begin{tabular}{lcccc}
\toprule
Agent & $\Delta BW\downarrow$ & $\sigma_{\Delta BW}\downarrow$ & $\Delta CF\downarrow$ & $\sigma_{\Delta CF}\downarrow$ \\
\midrule
SAA & 17.1366 & 14.8632 & 15.7886 & 16.0901 \\
PPO & 2.7254 & 5.1741 & \textbf{1.397}2 & \textbf{2.5187} \\
DQN & 0.2319 & 0.8640 & 9.5570 & 23.6005 \\
M-FOS & \textbf{0.2293} & \textbf{0.8542} & 1.6840 & 3.1335 \\
\bottomrule
\end{tabular}
\end{table}

\begin{table}[htbp]
\caption{Multi-Agent Total Rewards Across Frequency Bands (ARRC and simulation datasets)}
\label{tab:multiTotalRewards}
\centering
\small
\begin{tabular}{lccc}
\toprule
Agents & 2.4--2.5~GHz$\uparrow$ & 2.59--2.69~GHz$\uparrow$ & Simulation$\uparrow$ \\
\midrule
Random Static & -7{,}713 & -15{,}196 & -4{,}386 \\
SAA & -125{,}317 & -133{,}174 & -217{,}405 \\
\midrule
PPO & -6{,}603 & -10{,}690 & -27{,}498 \\
DQN & -11{,}894 & -14{,}001 & -6{,}495 \\
M-FOS & \textbf{-1{,}213} & \textbf{-9{,}599} & \textbf{-1{,}049} \\
\bottomrule
\end{tabular}
\end{table}

\section{Discussion}

\subsection{Overall Observations}

The primary objective of this work is not simply to improve single-agent spectrum access, but to investigate decentralized multi-agent cognitive radar spectrum sharing. Unlike conventional spectrum access methods that operate against a largely static environment, the proposed framework considers multiple adaptive radar systems simultaneously competing for shared spectrum without explicit communication or centralized coordination. The experimental results in Table~\ref{tab:multiResults} demonstrate that learning-based agents can successfully develop stable spectrum-sharing strategies under these constraints, with the proposed M-FOS framework generally providing the strongest overall performance across both recorded and simulated environments.

The performance improvement observed over the ablated M-FOS variant in most environments suggests that the meta-learning component contributes beyond the shared CNN-GRU policy architecture alone. The continuous-action formulation further allows finer adjustment of transmission center frequency and bandwidth than the discretized DQN action space, providing additional flexibility for balancing interference avoidance and waveform stability. Although DQN generally converged more rapidly because of its smaller exploration space, M-FOS typically learned policies that achieved higher long-term rewards in the decentralized multi-agent environments.

Most experiments in Table~\ref{tab:multiResults} evaluated three identical learning agents operating simultaneously within the same spectrum environment, allowing each algorithm to compete against copies of itself under identical conditions. Across these experiments, M-FOS achieved the highest average reward in the majority of recorded and simulated environments, outperforming SAA, PPO, and DQN in every case. The ablated M-FOS variant exceeded M-FOS in one environment in Table~\ref{tab:simMultiAgent}, suggesting that the evolutionary component is not universally advantageous under every spectrum condition. Additional experiments combining heterogeneous agent types (PPO, DQN, and M-FOS) with results shown in the bottom row of each table under Table~\ref{tab:multiResults} demonstrated that independently trained learning agents can also coexist effectively despite using different learning algorithms. The mixed agent experiments in Tables~\ref{tab:u245MultiAgent} and~\ref{tab:u264MultiAgent} even achieve better collective reward than the homogenous experiment with M-FOS. Collectively, these results suggest that decentralized reinforcement learning—and meta-learning in particular—provides an effective framework for long-term spectrum sharing among multiple cognitive radar systems.

\subsection{Single-Agent Performance}

The independent replay experiments in Table~\ref{tab:indResults} primarily serve as a baseline for comparison rather than the principal contribution of this work. Under these conditions, the heuristic Sense-and-Avoid (SAA) algorithm achieves strong reward performance because the most recent spectrum observation is often sufficiently representative of the immediately following transmission opportunity. Selecting the largest currently available contiguous region therefore minimizes collision penalties, which receive the largest weight in the proposed reward function.

The Random Static baseline also achieves reasonable performance despite having no adaptation capability. Because its transmission parameters remain fixed throughout an evaluation, it incurs no penalties associated with bandwidth or center-frequency distortion. Depending on the randomly selected frequency, this policy may either experience very little interference or remain permanently within congested spectrum. Averaging across multiple independent runs therefore provides a useful baseline while highlighting the limitations of completely non-adaptive spectrum access.

Although SAA minimizes immediate collisions effectively, it does so by continuously modifying its transmission bandwidth and center frequency. The resulting waveform distortion may be unacceptable for many coherent radar applications requiring stable transmission parameters over a CPI. Increasing the distortion penalties within the reward function could discourage this behavior; however, excessively weighting these penalties also suppresses exploration by the learning-based agents. Consequently, the proposed reward function intentionally balances collision avoidance with waveform stability rather than optimizing either objective independently.

\subsection{Multi-Agent Spectrum Sharing}

The primary contribution of this work lies in the proposed decentralized spectrum-sharing framework rather than simply improving the performance of an individual cognitive radar. Unlike traditional evaluations in which a single adaptive radar operates against a largely static RF environment, the proposed framework considers multiple independent cognitive radar systems simultaneously adapting their transmission strategies while competing for the same shared spectrum. Each radar selects its transmission center frequency and bandwidth independently using only its own local spectrum observations, with no communication, centralized coordination, or parameter sharing between agents. Consequently, all interaction occurs indirectly through the shared RF environment and the resulting reward signal.

This decentralized setting introduces temporal interactions that cannot be resolved through instantaneous spectrum sensing alone. Because spectrum ownership persists throughout an entire PRI rather than only during the transmitted pulse, an effective spectrum-sharing strategy must account for both current spectrum occupancy and future transmissions from competing radars. The learning problem therefore extends beyond selecting an immediately unoccupied frequency region to developing transmission behaviors that remain effective as multiple adaptive agents continuously modify the shared environment.

These characteristics expose an important limitation of the SAA algorithm. SAA assumes that the most recent spectrum observation accurately predicts future occupancy and therefore reacts only to instantaneous measurements. As a result, it frequently schedules transmissions between another radar's pulses while remaining unaware that the spectrum has effectively already been reserved for that radar's subsequent pulse. Although these transmissions may not overlap within a single spectrum snapshot, they nevertheless produce interference under the proposed spectrum-ownership model.

In contrast, the reinforcement-learning agents learn temporal occupancy patterns directly through repeated interaction with the environment. Rather than reacting only to the current spectrum snapshot, they learn policies that balance immediate spectrum availability with anticipated future occupancy while simultaneously considering bandwidth utilization and waveform stability. These capabilities become increasingly important as additional adaptive radar systems compete for the same RF resources.

The multi-agent experiments summarized in Table~\ref{tab:multiResults} demonstrate the effectiveness of the proposed decentralized learning framework across repeated heterogeneous and homogeneous multi-agent scenarios. In most evaluated environments, M-FOS achieved the strongest overall performance, while SAA's performance consistently degraded as additional adaptive radar systems shared the spectrum. These results suggest that learning-based agents are better able to balance long-term spectrum utilization, collision avoidance, and waveform stability than reactive heuristic approaches.

This trend becomes even more pronounced under higher agent densities. Table~\ref{tab:multiTotalRewards} summarizes a representative five-agent experiment in which all agents operated simultaneously within the same environment. Under these conditions, M-FOS achieved the highest cumulative reward in every evaluated environment, whereas SAA accumulated rewards more than an order of magnitude lower than the learning-based methods. Although extending SAA to incorporate longer observation histories could partially mitigate this weakness, such modifications would still not explicitly optimize long-term spectrum allocation, waveform stability, or adaptation to competing learning agents.

Taken together, these results suggest that the principal advantage of the proposed framework is not merely improved collision avoidance by an individual radar, but the emergence of effective decentralized spectrum-sharing strategies among multiple independently learning cognitive radar systems operating simultaneously without explicit coordination.

\subsection{Broader Discussion}

The experimental results demonstrate that the proposed framework provides a practical platform for studying decentralized cognitive radar spectrum sharing using both recorded over-the-air RF datasets and configurable simulated environments. Replay-based evaluation offers a repeatable benchmark for comparing competing algorithms while allowing identical spectrum conditions to be presented to every evaluated method. However, it does not capture fully closed-loop interactions between adaptive radar systems and external communication users. Consequently, the reported results should be viewed as an important intermediate step toward live over-the-air deployment rather than a complete operational validation.

The experiments focused on two measured frequency bands together with simulated environments designed to approximate similar occupancy characteristics. While these datasets provide diverse congestion levels and transmission patterns, evaluating additional frequency bands and operational scenarios would further demonstrate the generality of the proposed framework.

Similarly, the manually designed reward function successfully balanced transmission throughput, collision avoidance, and waveform stability across all evaluated environments. Future applications may benefit from adaptive reward formulations or multi-objective optimization techniques capable of automatically balancing competing operational objectives.

Finally, although the present work evaluates representative heuristic and reinforcement-learning approaches—including Random Static, SAA, PPO, DQN, M-FOS, and Ablated M-FOS—the proposed framework naturally extends to larger heterogeneous populations and additional learning architectures. Such studies may reveal increasingly complex cooperative and competitive behaviors as the number of simultaneously adapting radar systems grows.

\section{Conclusion and Future Work}

This paper presented a decentralized multi-agent framework for cognitive radar spectrum sharing in which multiple independent radar systems learn transmission strategies while operating within the same shared RF environment. This work contributes to the relatively limited literature on decentralized reinforcement learning for simultaneous cognitive-radar spectrum sharing to evaluate decentralized reinforcement learning for simultaneous cognitive radar spectrum sharing using multiple independently learning radar agents. Unlike conventional spectrum-sharing approaches that rely solely on instantaneous spectrum sensing, the proposed framework models spectrum access as a sequential multi-agent decision problem in which successful operation depends on learning temporal occupancy patterns and adapting to the behavior of competing radar systems.

Within this framework, a meta-learning extension of Model-Free Opponent Shaping (M-FOS) was developed by combining a shared CNN-GRU policy network with a genetic algorithm that adapts learning and exploration hyperparameters throughout training. Experimental evaluation using both recorded over-the-air datasets and configurable simulated environments demonstrated that M-FOS achieved the strongest overall performance across the majority of evaluated environments and produced the highest average reward overall while balancing collision avoidance, bandwidth utilization, and waveform stability.

Although the experiments were conducted using repeatable replay-based environments, the results demonstrate that decentralized reinforcement learning is capable of learning spectrum-sharing strategies that account for temporal interactions among independently operating cognitive radar systems. While heuristic methods such as Sense-and-Avoid perform well in isolated replay scenarios, they become increasingly limited when multiple adaptive radar systems simultaneously compete for spectrum. In contrast, the proposed learning framework enables agents to implicitly develop transmission strategies that remain effective under dynamic multi-agent interactions.

Future work will investigate larger heterogeneous radar populations, alternative meta-learning strategies beyond genetic algorithms, transformer-based temporal policy architectures, adaptive reward design, and fully closed-loop over-the-air experimentation involving multiple simultaneously operating cognitive radar systems. Extending the framework to cooperative learning, communication-assisted spectrum sharing, and mixed radar-communication coexistence scenarios also represents promising directions for future research.

\section*{Acknowledgment}
The authors would like to thank Dr. Shane Flandermeyer for providing feedback throughout this work and supplying the ARRC spectrum datasets used during experimentation, Department 70 at Tinker Air Force Base for assisting in resources for data collection, and the Department of War Science, Mathematics, and Research for Transformation (DoW SMART) Scholarship-for-Service Program for funding this work.

The views expressed in this article are those of the authors and do not reflect the official policy or position of the United States Air Force, Department of Defense, or U.S. Government.

{\appendices
\section{Hyperparameters and Experimental Setup}
To improve reproducibility, all experiments were conducted using fixed random seeds. Multi-agent experiments were executed concurrently within the same shared environment.

\subsection{Environment and Signal Processing Parameters}

The radar environment used an FFT size of 1024 frequency bins over a 100~MHz channel bandwidth, corresponding to a frequency-bin resolution of approximately 0.0977~MHz per bin. The PRI was set to 204.8 $\mu$sec with a CPI length of 256 pulses. 

Each agent was initialized with randomized starting indices within the pulse interval to reduce synchronization bias and encourage diverse interaction patterns.

\subsection{PPO Hyperparameters}

The PPO agent hyperparameters are shown in Table~\ref{tab:ppo_hyperparameters}.

\begin{table}[h]
\centering
\caption{PPO Hyperparameters}
\begin{tabular}{|l|c|}
\hline
\textbf{Parameter} & \textbf{Value} \\ \hline
Time horizon & 1024 \\ \hline
Discount factor ($\gamma$) & 0.95 \\ \hline
GAE lambda & 0.95 \\ \hline
Clipping parameter ($\epsilon$) & 0.2 \\ \hline
Optimization epochs & 10 \\ \hline
Learning rate & $5 \times 10^{-4}$ \\ \hline
Batch size & 16 \\ \hline
Truncated BPTT length & 32 \\ \hline
Entropy coefficient (initial) & $1 \times 10^{-3}$ \\ \hline
Entropy decay rate & 0.995 \\ \hline
Minimum entropy coefficient & $2 \times 10^{-3}$ \\ \hline
\end{tabular}
\label{tab:ppo_hyperparameters}
\end{table}

\subsection{DQN Hyperparameters}

The DQN agent operated with a discrete action space generated from bandwidth–center frequency combinations. Candidate bandwidths were $\{96, 128, 160\}$ FFT bins with 32 uniformly spaced center frequencies across the FFT spectrum.

The DQN hyperparameters are shown in Table~\ref{tab:dqn_hyperparameters}.

\begin{table}[h]
\centering
\caption{DQN Hyperparameters}
\begin{tabular}{|l|c|}
\hline
\textbf{Parameter} & \textbf{Value} \\ \hline
Replay memory size & 2000 \\ \hline
Batch size & 32 \\ \hline
Discount factor ($\gamma$) & 0.9 \\ \hline
Learning rate & $1 \times 10^{-3}$ \\ \hline
Target network update interval & 1000 pulses \\ \hline
Hidden layer sizes & [512, 256, 84] \\ \hline
Initial exploration rate ($\epsilon$) & 1.0 \\ \hline
Minimum exploration rate ($\epsilon_{\min}$) & 0.05 \\ \hline
Exploration decay rate & 0.9995 \\ \hline
\end{tabular}
\label{tab:dqn_hyperparameters}
\end{table}

\subsection{M-FOS Hyperparameters}

The M-FOS agent employed an evolutionary optimization strategy with randomly initialized genomes. Each genome was comprised of a learning rate $\alpha$, discount factor $\gamma$, exploration center and exploration bandwidth, and entropy coefficient. The parameters used for the GA are shown in table~\ref{tab:mfos_hyperparameters}.

\begin{table}[h]
\centering
\caption{M-FOS Hyperparameters}
\begin{tabular}{|l|c|}
\hline
\textbf{Parameter} & \textbf{Value} \\ \hline
Population size & 5 \\ \hline
Mutation scale & 0.05 \\ \hline
Elite fraction & 0.4 \\ \hline
Fresh population fraction & 0.2 \\ \hline
\end{tabular}
\label{tab:mfos_hyperparameters}
\end{table}

\subsection{Reward Function Parameters}

The reward function sought to prioritize collision avoidance first, and distortion avoidance second with the selected weights for the experiments shown in this paper shown in table~\ref{tab:reward_parameters}.

\begin{table}[h]
\centering
\caption{Reward Function Parameters}
\begin{tabular}{|l|c|}
\hline
\textbf{Parameter} & \textbf{Value} \\ \hline
Clean transmission weight & 1.0 \\ \hline
Collision penalty weight & 30.3 \\ \hline
Bandwidth distortion penalty weight & 0.5 \\ \hline
Center-frequency distortion penalty weight & 0.5 \\ \hline
\end{tabular}
\label{tab:reward_parameters}
\end{table}

}

\printbibliography

\begin{IEEEbiography}
[{\includegraphics[width=1in,height=1.25in,clip,keepaspectratio]{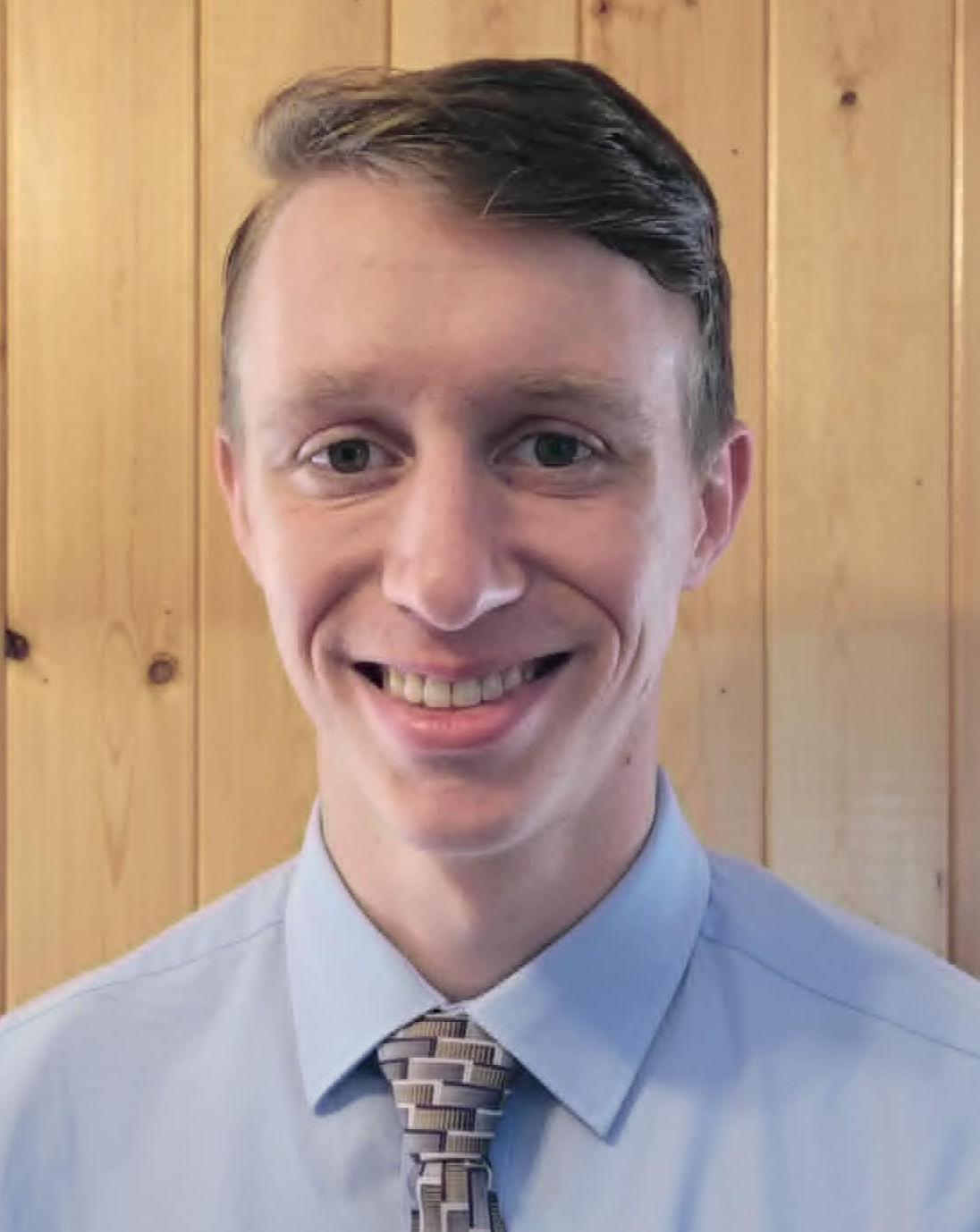}}]{Job Elliott (Graduate Student Member, IEEE)}
Received the Associate's degree in intelligence studies from the Community College of the Air Force in 2018, the bachelor's degree in computer science from the University of Central Oklahoma in 2020, and the master's degree in computer science with an emphasis in artificial intelligence from Oklahoma Christian University in 2023. He is currently pursuing his Ph.D. in computer science at the University of Oklahoma.

He previously served in intelligence and geospatial analysis roles within the Air National Guard, and he has worked for the air force since 2021 as a civilian programmer.  

His research interests include reinforcement learning, deep learning, cognitive radar, and radar resource management. He is a recipient of the DoW SMART Scholarship.

\end{IEEEbiography}

\begin{IEEEbiography}
[{\includegraphics[width=1in,height=1.25in,clip,keepaspectratio]{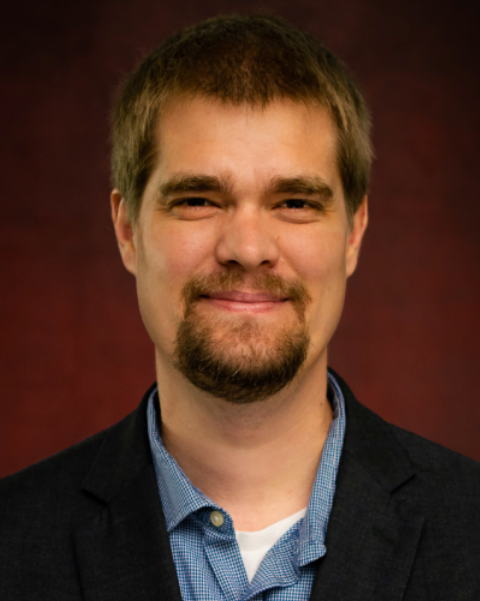}}]{Justin G. Metcalf (Senior Member, IEEE)}
received the bachelor’s degree in computer engineering from Kansas State University, Manhattan, KS, USA, in 2006, and the master’s and Ph.D. degrees (Hons.) in electrical engineering from The University of Kansas, Lawrence, KS, USA, in 2011 and 2015, respectively. 

He was a KSU Presidential Scholar with Kansas State University. From 2006 to 2008, he was with Flight Simulation Labs, Lockheed Martin Aeronautics, Fort Worth, TX, USA. From 2008 to 2014, he was with the Radar Systems Laboratory, The University of Kansas. From 2014 to 2018, he was a Research Electronics Engineer with the Sensors Directorate of the Air Force Research Laboratory, Wright-Patterson AFB. Since 2018, he has been an Assistant Professor with the Department of Electrical and Computer Engineering, The University of Oklahoma, Norman, OK, USA, and a member of the Advanced Radar Research Center. 

Dr. Metcalf is a member of the IEEE AESS Radar Systems Panel. He was a recipient of the Richard and Wilma Moore Award for the Best Departmental Master’s Thesis from 2011 to 2012, the 2017 IEEE Dayton Section Young Professionals Award, the 2020 DARPA Young Faculty Award, and the 2023 IEEE AESS Fred Nathanson Memorial Radar Award. From 2016 to 2018, he was the Chair of the Dayton Chapter of the IEEE Aerospace and Electronic Systems Society. 

\end{IEEEbiography}

\begin{IEEEbiography}
[{\includegraphics[width=1in,height=1.25in,clip,keepaspectratio]{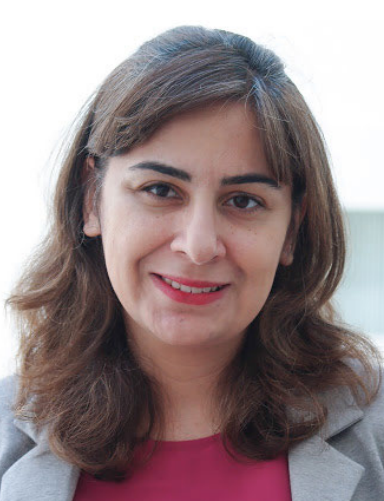}}]{Golnaz Habibi (Senior Member, IEEE)}
received the B.Sc. degree in electrical and control engineering from the K. N. Toosi University of Technology, Iran, in 2005, the M.Sc. degree in control engineering from Tarbiat Modares University, Iran, in 2007, and the Ph.D. degree in computer science from Rice University, in 2015. She is currently a Research Scientist with the Department of Aeronautics and Astronautics, MIT. Her research interests include robotics, control systems, machine learning, multi agent systems, visual navigation, reliable communication, and improving the safety and reliability of autonomous agents. Her paper has been nominated for best student paper award in DARS 2012 and she received the K2I award by Chevron Corporation, in 2013.

\end{IEEEbiography}

\end{document}